\documentstyle[epsf,a4]{article}

\def\jape{{\sf J\kern-.15em\raise-.3ex\hbox{\footnotesize\sf
A}\kern-.15em\raise.5ex\hbox{\footnotesize\sf
P}\kern-.15em\kern.120emE\kern.120em-\kern.120em1}}

\title{A symbolic description of punning riddles
and its computer implementation\thanks{A talk on this work was given
at the International Conference on Humor and Laughter, Luxembourg, 1993,
and a short version will appear in the Proceedings of the Twelfth
National Conference on Artificial Intelligence (AAAI-94), Seattle, USA,
August 1994. 
} 
} 

\author{Kim Binsted\\
Graeme Ritchie\\
Department of Artificial Intelligence\\
University of Edinburgh\\
Edinburgh EH1 1HN\\
Scotland.
}

\date{ }       

\begin{document}
\epsfverbosetrue

\maketitle

\begin{abstract}
Riddles based on simple puns can be classified according to the
patterns of word, syllable or phrase similarity they depend upon.
We have devised a formal model of the semantic and syntactic regularities
underlying some of the simpler types of punning riddle. We have also
implemented this preliminary theory in a computer program which
can generate riddles from a lexicon containing general data about
words and phrases; that is, the lexicon content is not customised
to produce jokes.
Informal evaluation of the program's results by a set of human
judges suggest that the riddles produced by this program are
of comparable quality to those in general circulation among
school children.

\end{abstract}

\section{Introduction}
\label{introduction}

One very common form of humour is the question-answer joke, or riddle. A
significant subset, probably the vast majority, of these jokes are based
on  some form of pun (question-answer puns make up almost a
third of the riddles in \cite{CAJB}). For example:

\begin{verse}

What do you get when you cross a sheep with a kangaroo?\\
{\it A woolly jumper}

\end{verse}

Riddles of this general sort are of particular interest for a number of
reasons. The linguistics of riddles has been investigated before (e.g.
\cite{PG}). Also, there is a large corpus of riddles to examine: books
such as the Crack-a-Joke Book \cite{CAJB} record them by the thousand.
Finally, riddles exhibit regular structures and mechanisms, which could
be modelled and used to generate new riddles.

There appear to certain well-defined subclasses of this genre,
distinguished by the type of pun, or the arrangement of punning words
and phrases. We have devised a formal model of the punning mechanisms
underlying some of these subclasses; i.e.\ a symbolic account of the
configurations of lexical entities required to make up each of these
types of riddle.  In addition, we have implemented a computer program
which uses these symbolic rules to construct punning riddles from a
lexicon (i.e.\ a store of words or phrases with syntactic and semantic
information about each entry), and have used this system to generate a
large number of jokes, including some which were unknown to the authors
(and are perhaps novel).  An informal evaluation of the performance of
this program suggests that its output is not significantly worse than
that produced by human composers of such riddles.

In Section \ref{methodological-issues} we will set out the assumptions
of our approach, which relies more on generative linguistics and
artificial intelligence than on literary or psychological studies. Then
we shall outline the scope of our project (Section
\ref{jokes-tackled}), introduce the theoretical constructs we use to
describe punning riddles (Section \ref{theory}), and the computer
program (\jape) which embodied these ideas. After summarising the
evaluation which we carried out (Sections  \ref{evaluation} and
\ref{results}), we shall
relate the work to other research (Section \ref{others}), and suggest
some ways in which these ideas could be developed further (Section
\ref{further-work}).

\section{Methodological issues}
\label{methodological-issues}

\subsection{Generative linguistics}

One of the major contributions which Chomsky \cite{chomsky57,chomsky65}
made to the field of
theoretical linguistics was the establishment of a methodological
framework for the study of language.
There were several tenets to this approach, either explicitly or
implicitly:
\begin{enumerate}
\item
The aim is to define symbolic rules and structures  which characterise
what constitutes a sentence of a language and what does not.
\item
These descriptions should be sufficiently precise and detailed
that there is no doubt about what they predict, to the extent
that it would {\em in principle} be possible to check the
predictions mechanically (e.g.\ using a computer).
\item
These symbolic accounts can be empirical without large scale
collection of data or statistical studies. The linguist compares
what sentences are known to exist in the language with those
predicted by the rules. Often, falsification of a proposed set of
rules can be achieved from a relatively small set of examples.
\item
The rules should capture regularities in the data. That is, sentences
which have some inherent similarity (syntactically, semantically, etc.)
should be described in a similar way by the rules, and systematic
alterations to the symbolic descriptions should correspond to
systematic changes in the language phenomena being described.

\end{enumerate}

We have to a large extent adopted these attitudes in our study of riddles,
as have various other humour researchers (some tacitly). We have
attempted to devise abstract symbolic accounts of the detailed
mechanisms underlying our chosen set of phenomena (certain types of
punning riddle), we have defined these rules precisely (as shown by the
computer implementation), and we believe that they show regularities in
exactly the way that linguists expect grammars to display
generalisations about sentences.

\subsection{Artificial intelligence}

Within artificial intelligence, a research paradigm sometimes known as
{\em experimental programming} is common. In this, a computer program is
used to explore ideas, rather than to provide a final and conclusive
test of a single well-articulated theory or to operate as a polished
piece of software.  Although the methodology owes a lot to the notion of
an ``experiment'' in traditional science, it does not rely on critical
experiments to falsify abstract theories (although it would be
satisfying if this were the case). Rather, the researcher attempts to
clarify his/her ideas by posing the question: ``what would it take to
have a computer program perform this task?''.  Concrete fleshing out of
embryonic ideas then consists of trying to construct a computer program.
This not only forces a degree of detail and precision (cf.\ comments
above on linguistic methodology), it also provides a readily testable
version of the draft theory.  Running the program and observing its
behaviour (not just its final results but also how it achieves them) can
provide useful insights into weaknesses of the ideas, and may even
inspire possible amendments or extensions to the proto-theory. (See
\cite{buchanan,newell,ritchie} for further discussion of this
approach).

The work described here can be seen as exemplifying this approach.
The central task of the project was to design, implement, and test a
computer program, but the important product of the work was {\em not}
the program itself; rather, it was the set of ideas that we developed
in the course of the work.

\subsection{Humour and artificial intelligence}

Artificial intelligence (AI) has delved into so many areas of human
behaviour --- vision, mathematics, planning and natural language
processing, among others --- that it is sometimes difficult to find an
area which is {\em not} within AI's research domain. If, as Minsky
claims, a suitable goal for AI research is to get a computer to do
``\ldots a task which, if done by a human, requires intelligence to
perform,'' \cite{Minsky2} then most of human experience is a fit subject
for research.

For theoretical research, however, there is also the requirement that
the results must be {\em falsifiable} --- that is, it must be possible
to do an experiment which could disprove any claims of success. This
makes the artistic side of human nature hard to investigate because,
although art is definitely within the domain claimed by AI, it is
difficult (if not impossible) to disprove the claim  ``this is good
art'' (or even ``this is art'') no matter who or what produced the work
in question. Poetry, painting, and music all suffer from this problem.

A second factor to take into account is whether or not a formal model of
the task has been, or could feasibly be, devised. For a task to be
computationally tractable, it is necessary (but not sufficient) that a
formal description of the task can be developed.

These two constraints, falsibility and formalisability, reduce the
domain of AI research considerably, particularly when looking at
artistic intelligent behaviour.

Humour generation is falsifiable in that there is a simple test of its
success: whether the audience is amused or not.  Subject to various
caveats, this gives us a relatively rigorous way of testing results.

Although no computationally tractable model of humour as a whole has yet
been developed, we believe that by tackling a limited and
linguistically-based set of phenomena, it is realistic to develop a
formal symbolic account.

\section{Phenomena considered}
\label{jokes-tackled}

\subsection{Pepicello and Green's riddle theory}

There are numerous collections and analyses of riddles, from the
viewpoints of anthropology, sociology, literature and related fields.
Apart from providing an overview of humour (and possibly some goal jokes
to replicate), however, these works were too discursive or informal to
be of direct use in this project. In fact, the only work that explores
this genre of joke in enough detail is \cite{PG}.

In their book, Pepicello and Green describe the various grammatical,
written and visual strategies incorporated in riddles\footnote{This
subsection is essentially a precis of chapters two and three of
\cite{PG}, and the examples are theirs.}. They
hold the common view that humour is closely related to ambiguity,
whether it be linguistic (such as the phonological ambiguity in a
punning riddle) or contextual (such as riddles that manipulate social
conventions to confuse the listener). Moreover, they claim that humour
depends on that ambiguity being `unsolvable' by the listener, at least
until the punchline resolves it in some unexpected way.

Linguistic ambiguity can take place at the phonological, morphological,
or syntactic levels of grammar. For example, the sentence
``John lives near the bank'' is
phonologically ambiguous, since the noun ``bank'' can refer to either a
building where money is stored,  or the shore of a river. The sentences
``The book is read,'' and ``The book is red'', however, are
morphologically ambiguous, since ``read'' is only phonetically identical
with ``red'' in its past participle form. Finally, the sentence ``John
looked over the car'' is syntactically ambiguous, since it has two
distinct grammatical analyses.

Each kind of ambiguity, or a combination, can be used in riddles. For
example:
\begin{enumerate}
\item {\bf Phonological:} What bird is lowest in spirits? {\em A bluebird.}
\item {\bf Morphological:} Why is coffee like soil? {\em It is ground.}
\item {\bf Syntactic:} Would you rather have an elephant kill you or a
gorilla? {\em I'd rather have the elephant kill the gorilla.}
\end{enumerate}

As can be seen from these examples, the ambiguity can occur either in
the question (3) or the punchline (1 and 2).

Pepicello and Green go on to describe many different strategies used in
riddles to produce and manipulate these linguistic ambiguities. However,
what all these strategies have in common is that they ask the `riddlee'
to accept a similarity on a phonological, morphological, or syntactic
level as a point of {\em semantic} comparison, and thus get fooled. For
example, the riddle:

\begin{quote}

Why is a river lazy? {\em Because it seldom gets out of its bed.}
\cite{CAJB}

\end{quote}
uses the phonological ambiguity in the word ``bed'' to imply that a
river bed is semantically identical with a sleeping bed, and therefore
that not getting out of a river bed is a sign of laziness.

Using this analysis, we can choose a subset of riddles which share {\em
underlying properties} (e.g.\ type of ambiguity, strategy for deceiving
the riddlee), and which are thus able to be generated by closely related
mechanisms. This is in contrast to jokes which have in common merely
their {\em surface form} (e.g.\ jokes that ridicule a stereotyped group
by describing how they might go about screwing in a lightbulb.)

\subsection{A source of data}

Most of the riddles discussed below come from ``The Crack-a-Joke Book''
\cite{CAJB}, a collection of jokes chosen by British children. This is
an ideal source for several reasons. The riddles are simple, requiring
only basic English to understand; their humour generally arises from
their punning nature, rather than their subject matter (children do not
seem eager to joke about God, politics or taxes, and the Crack-a-Joke
Book was undoubtably edited for sex and toilet humour); and there are a
huge number of riddles to to choose from.
For much the same reasons, however, The Crack-a-Joke Book riddles are
unlikely to make an adult reader laugh. This is unfortunate, but
unavoidable --- it would be over-ambitious to tackle sophisticated adult
humour at this stage.

The Crack-a-Joke Book includes some non-riddle humour, as well:
limericks, dialogues, and punning book titles, for example. None of
these types will be discussed below, as they are outside the range of
this project.

Some of the jokes in the corpus come from sources other than the
Crack-a-Joke book. If a given riddle comes from a book, it is cited; if
there is no citation, then the riddle was remembered from casual
conversation.

\subsection{Punning riddles}

Riddles can be divided into groups in a variety ways --- by subject
matter, narrative strategy, source, etc.  We decided to select a subset
of riddles which exploited linguistic ambiguity in a similar way.
Although there are
riddles that use two or more types of ambiguity, it is relatively
straightforward to divide the bulk of question-answer riddles according
to the primary level of ambiguity they use.
Most of the riddles in \cite{CAJB} use phonological,
morphological, or syntactic ambiguity.  Of these three types,
phonologically ambiguous riddles seem most directly amenable to
formalisation in a computationally tractable manner. From now on,
riddles of this type will be called {\em puns}\footnote{Note that other
works may use the word ``pun'' differently.} or {\em punning riddles}.

\label{tax}

There are three main strategies used in puns to exploit phonological
ambiguity: {\em syllable substitution}, {\em word substitution}, and
{\em metathesis}. This is not to say that other strategies do not exist;
however, none were found among the large number of punning jokes
examined.

\subsubsection{Syllable substitution}

Puns using this strategy confuse a syllable (or syllables) in a word
with a similar- or identical-sounding word. For example:

\begin{quote}
What do short-sighted ghosts wear? {\em Spooktacles.} \cite{CAJB}
\end{quote}

The word containing the ambiguous syllable usually appears in the
punchline of the riddle, while the question of the riddle refers to
some constructed `meaning' (i.e.\ not the real meaning) of the word.
However, the reverse can also occur:

\begin{quote}
What is an octopus? {\em An eight-sided cat.} \cite{CAJB}
\end{quote}

Sometimes, several word-syllable confusions are made:

\begin{quote}
Where are whales weighed? {\em In a whaleweigh station.} \cite{CAJB}
\end{quote}

Note that sometimes the confused syllable is actually {\em replaced}
with a similar-sounding word (e.g.\ in ``spooktacles'' and ``whaleweigh''),
whereas other times the possible substitution is just referred to (e.g.
``octopus'' was not changed to ``octopuss''), probably because the
appearance of the changed word in the question would give the joke
away.

Pepicello and Green have argued that some riddles of this type
are in fact {\em morphologically} ambiguous, since a whole word is being
confused with a morpheme (e.g.\ the ``pus'' suffix in ``octopus'').
However, jokes that use syllable substitution are very similar in
structure, whether the syllable being confused is a morpheme or not, so
we will regard them as essentially the same.

\subsubsection{Word substitution}

Word substitution is very similar to syllable substitution. In this
strategy, an entire word is confused with another similar- or
identical-sounding word. For example:

\begin{quote}
How do you make gold soup? {\em Put fourteen carrots in it.} \cite{CAJB}
\end{quote}

Again, the confused word can appear in the question instead of the
punchline:

\begin{quote}
What do you do if you find a blue banana? {\em Try to cheer it up.}
\cite{CAJB}
\end{quote}

Note that in the above joke, ``blue'' has two meanings, but only one
surface form. This demonstrates that a word can be confused with: an
alternate meaning (e.g.\ ``blue'', the colour, with ``blue'', the mood); a
word spelled differently but sounding the same (e.g.\ ``carats'' with
``carrots''); or a word that sounds slightly different, as in:

\begin{quote}
Where do elves go to get fit? {\em Elf farms.} \cite{CAJB}
\end{quote}

The confused word is often part of a common phrase. For example, in:

\begin{quote}
What sits in a fruit bowl and shouts for help? {\em A damson in
distress.} \cite{CAJB}
\end{quote}
the joke relies on the riddlee recognizing the punchline as a warped
version of the phrase ``damsel in distress.''

\subsubsection{Metathesis}

Metathesis is quite different from syllable or word substitution. Also
known as {\em spoonerism}, it uses a reversal of sounds and words to
suggest (wrongly) a similarity in meaning between two
semantically-distinct phrases. For example, this joke:

\begin{quote}

What's the difference between a torn flag and a bent sixpence? {\em
One's a tattered banner and the other's a battered tanner.}

\end{quote}
implies that tattered banners and battered tanners actually have a lot
in common, the only difference being the reversed initial letters.\footnote{
A ``sixpence'' is an obsolete British coin, and ``tanner'' was a slang
name for that coin. We apologise for the age and cultural bias of some
of our examples.}

Words, as well as letters, can be reversed, for example:
\begin{quote}
What is the difference between a donkey and a postage stamp? {\em One
you lick with a stick, the other you stick with a lick.} \cite{PG}

\end{quote}
Again, there is the implied similarity between donkeys and stamps, just
because some of the actions you can perform on them are phonologically
similar.

\subsection{Chosen class of riddles}

All three of the above types of pun are potentially tractable
for detailed formalisation and hence computer generation.  We chose to
concentrate on word-substitution puns, simply because lists of
phonologically identical words ({\em homophones}) are readily available,
whereas the other two types require some kind of sub-word comparison.

In particular, the class of jokes on which we have concentrated all:
\begin{itemize}
\item have the substituted word in the {\em punchline} of the joke,
rather than the question, and

\item substitute a homophone for a word in a {\em common noun phrase}.
\end{itemize}

These restrictions are simply to reduce the scope of the research even
further, so that the chosen subset of jokes (punning riddles with
noun-phrase punchlines) can be covered in a comprehensive, rigorous
manner. The model and method used to generate such jokes, however,
should be general enough that it can be expanded in a straightforward
way to cover all word-substitution puns, and --- eventually --- sub-word
puns and spoonerisms.

\section{Symbolic descriptions}
\label{theory}

\subsection{Introduction}

Now that we have restricted the goal set of jokes to simple
question-answer puns with noun-phrase punchlines,
it is necessary to describe the common features of their structure in
such a way that they could be constructed by a computer.

These features must be specified well enough that any piece of text that
has them is a joke. It is {\em not} good enough to say that there are
some jokes among the pieces of text that share these features. In
keeping with the remarks made about linguistic methodology in
Section~\ref{methodological-issues}, we are trying to devise rules which
will define as exactly as possible our chosen class of jokes, without
some further intervention by human `intuition' or `commonsense'.
Although we expect that one way of extending our theory would be to
state evaluative rules (heuristics) which rated the quality of jokes,
we have no such mechanisms at present.

On the other hand, not all jokes (nor even all question-answer punning
riddles) need have the stated features --- to completely specify everything
that could make text funny is too ambitious a task at present.

\subsection{A model of simple question-answer riddles}
\label{model}

As mentioned in section~\ref{jokes-tackled}, one reason riddles can be
funny is that they confuse linguistic levels. There are several common
ways of taking advantage of this confusion of levels. One method
(word--word or word--syllable substitution) is to take a valid English
word or phrase, and substitute into it a word that is phonologically
similar to part of the word/phrase. The riddle then uses resulting
nonsense word or phrase as if it were a semantically sensible
construction, just because it is phonologically similar to a real word
or phrase. The effective meaning of this fake construction is a
combination of the meanings of the two pieces of text used to build it.
For example:

\begin{quote}
What do you give an elephant that's exhausted? {\em Trunkquillizers.}
\cite{CAJB}

\end{quote}
In this joke, the word ``trunk'', which is phonologically similar to the
syllable ``tranq'', is substituted into the valid English word
``tranquillizer''. The resulting fake word ``trunkquillizer'' is given a
meaning, referred to in the question part of the riddle, which is some
combination of the meanings of ``trunk'' and ``tranquillizer'':

\begin{description}
\item[trunkquillizer:] A tranquillizer for elephants.
\end{description}

Note that this is not the only meaning for ``trunkquillizer'' that could
produce valid jokes. For example:

\begin{quote}
What kind of medicine gives you a long nose? {\em Trunkquillizers.}
\end{quote}
is a joke (if not a good one) based on a different definition for
``trunkquillizer'' --- namely, `a medicine that gives you a trunk'. The
fake meaning should combine notable semantic features of both of the
valid words/phrases used to construct it, so that the riddle question
will be a reasonable description of the nonsense concept.

In non-humorous communication, we can use the meaning of a real
word/phrase to build a question that has the word/phrase as an answer:

\begin{quote}

What do you call someone who douses flames? {\em A firefighter.}

\end{quote}
Similarly, riddles like the `trunkquillizer' example use the constructed
meaning of a fake word/phrase to build a question that {\em would} have
the word/phrase as an answer, if it really existed. This question
becomes the first part of the riddle, and the fake word/phrase becomes
the punchline.

The building-blocks we have identified so far, are:
\begin{enumerate}
\item a valid English word/phrase
\item the meaning of the word/phrase
\item a shorter word, phonologically similar to part of the word/phrase
\item the meaning of the shorter word
\item a fake word/phrase, made by substituting the shorter
word into the word/phrase
\item the meaning of the fake word/phrase, made by combining the
meanings of the original word/phrase and the shorter word.
\end{enumerate}

\begin{figure}[htb]
\epsfxsize = 0.75\textwidth
\epsfbox{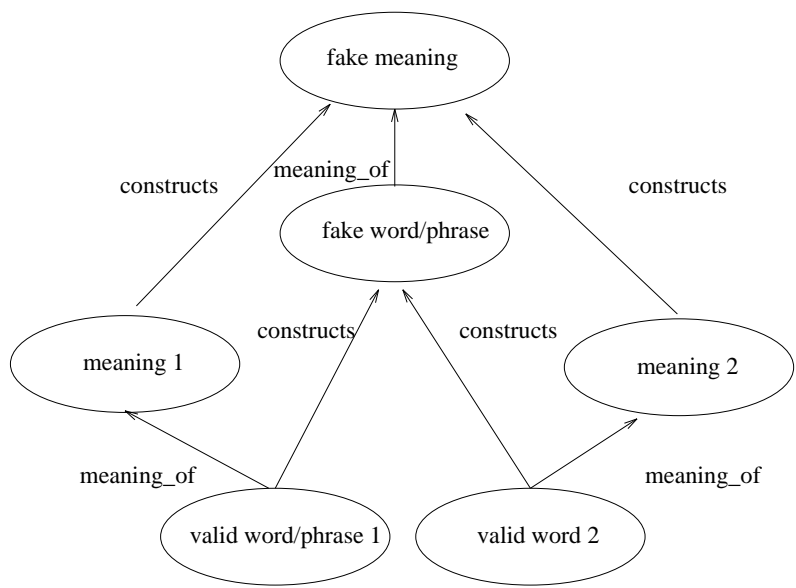}
\caption{The relationships between parts of a pun}
\label{theoryfig}
\end{figure}

The relationships between these building blocks are shown in
figure~\ref{theoryfig}.

At this point, it is important to distinguish between the mechanism for
building the {\em meaning} of the fake word/phrase, and the mechanism
that uses that meaning to build a question with the word/phrase as an
answer. For example, the following questions use the same meaning for
`trunkquillizer', but refer to that meaning in different ways:

\begin{itemize}
\item What do you use to sedate an elephant?
\item What do you call elephant sedatives?
\item What kind of medicine do you give to a stressed-out elephant?
\end{itemize}
On the other hand, {\em these} questions are all put together in the
same way, but from different fake meanings:

\begin{itemize}
\item What do you use to sedate an elephant?
\item What do you use to sedate a piece of luggage?
\item What do you use to medicate a nose?
\end{itemize}

We have adopted the term {\em schema} for the first sort of symbolic
description (i.e.\ the underlying configuration of meanings and words),
and {\em template} for the second sort (i.e.\ textual rules or patterns
used to construct an appropriate question-answer pair).

\subsection{Lexicon}

Before going into more detail about schemata and templates, we should
first summarise what we have to assume about the lexicon (dictionary)
used by these rules.

There is a (finite) set of {\em lexemes}. A lexeme is an abstract
entity, roughly corresponding to a meaning of a word or phrase. Each
lexeme has exactly one entry in the lexicon, so if a word has two
meanings, it will have two corresponding lexemes. Each lexeme may have
some {\em properties} which are true of it (e.g.\ being a noun), and
there are a number of possible {\em relations} which may hold between
lexemes (e.g.\ synonym,  homophone, subclass). Each lexeme is also
associated with a {\em near-surface form} which indicates (roughly) the
written form of the word or phrase.
We will deliberately blur the distinction between relations which are
explicitly represented in the lexicon and other inter-lexeme relations
which could be defined from the actual concrete entries. For example,
one could define a relationship ``superordinate-synonym'' such
that X is a superordinate-synonym of Y if there is a lexeme Z
which is a superordinate of Y and Z is a synonym of X.
We shall use the term ``lexical relation'' to denote any relation
between lexemes, regardless of whether a given lexicon might store
it explicitly.

The implemented computer program (see Section \ref{program}) represented
the lexicon as fairly conventional data structures, but the details of
those mechanisms are not part of the theory. In particular, the {\em
homophone} relation was represented as a separate data structure, the
{\em homophone base}, based on data collected from an independent source.

\subsection{Schemata}

A {\em schema} stipulates a set of relationships which must hold
between the lexemes used to
build a joke. More specifically, a schema determines how real
words/phrases are glued together to make a fake word/phrase, and which
parts of the lexical entries for real words/phrases are used to
construct the meaning of the fake word/phrase.
Strictly speaking, the relationships are between {\em sequences}
of lexemes, but in most cases the sequence will be of length one
(a single lexeme); there are only a few examples so far where
we need the flexibility to link to a longer sequence of lexemes.

\begin{figure}[htb]
\epsfxsize = 0.75\textwidth
\epsfbox{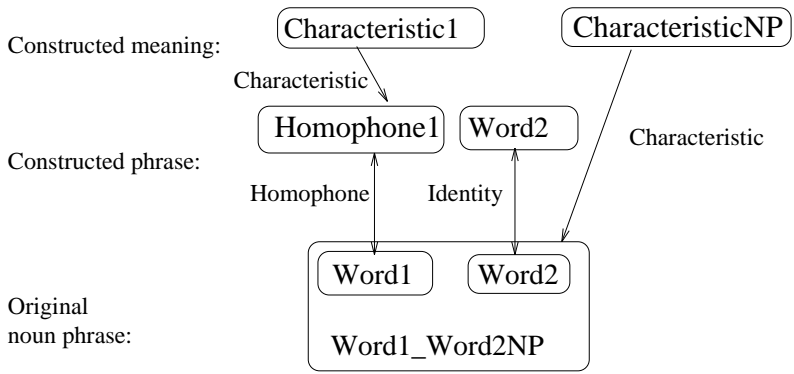}
\caption{The {\em lotus} schema}
\label{lotusfig}
\end{figure}

There are many different possible schemata\footnote{The various
schemata acquired obscure labels, with little mnemonic value, in the
course of the project; these have no particular significance.}. For
example, the schema in figure~\ref{lotusfig} constructs a fake phrase by
substituting a homophone for the first word in a real phrase, then builds
its meaning from the meaning of the homonym and the real phrase.
Another schema might construct a fake phrase by substituting a homophone
for the {\em second} word in a real phrase, then constructing its
meaning from the {\em first} word in the phrase and the homophone. (This
is, in fact, exactly what the {\em jumper} schema does.)

The schema shown in figure~\ref{lotusfig} is {\em uninstantiated};
that is, the actual lexemes to use have not yet been specified.
Moreover, the {\em possible} relationships between the lexemes are
stated, but the exact choice of relations is still unspecified.
Instantiating a schema means choosing lexemes to go in the schema, and
specifying the exact relationships between those lexemes.

More formally, an uninstantiated schema is a set of variables and a
set of constraints (links) between those variables.  Constraints
indicate relations that must hold between any values assigned to those
variables.  For a schema to be instantiated, each variable
must be instantiated (or ``bound'') to a (sequence of) particular
lexeme(s) satisfying the those constraints. (As noted earlier, in the
typical case the sequence will be a single lexeme, but in general it
could be several lexemes, to represent a phrase.)  Each constraint
(link) in the schema stipulates some {\em set} of possible relations
that must hold between the lexemes it connects. This stipulation may
be very narrow (just one relation) or very wide (any relation at all).
In our initial implementation, we have found a need for only three
types of link:

\begin{description}

\item[Homophone:]  The linked lexemes are homophones
(whether spelt the same or not), as specified in the homophone base
(see section~\ref{homonyms}).

\item[Identity:] The linked lexemes are identical, and
therefore have exactly the same entry in the lexicon.

\item[Characteristic:] There is some relationship between the
lexemes, defined in terms of (but not necessarily explicitly
represented in) their lexical entries.

\end{description}

For example, in the lexicon, the lexeme {\bf
spring\_cabbage} might participate in relations  as in Table
\ref{spring-cabbage}.
If {\bf spring\_cabbage} were to be included in a schema, at one end of
a {\em characteristic} link, the other end of the link could be
associated with any of these values ({\em vegetable,
garden}, etc), depending on the exact label ({\em class, location}, etc.)
chosen for the {\em characteristic} link. When the schema is
fully instantiated, however, both the characteristic link and the
lexemes it joins are specified.

\begin{table}
\begin{tabular}{|l|l|}
\hline
RELATION & VALUE \\
\hline
class & vegetable \\
location & garden \\
action & grows \\
adjective & green \\
etc\ldots & \ldots \\
\hline
\end{tabular}
\caption{\label{spring-cabbage} Example of Lexical Relations}

\end{table}

\begin{figure}[htb]
\epsfxsize = 0.75\textwidth
\epsfbox{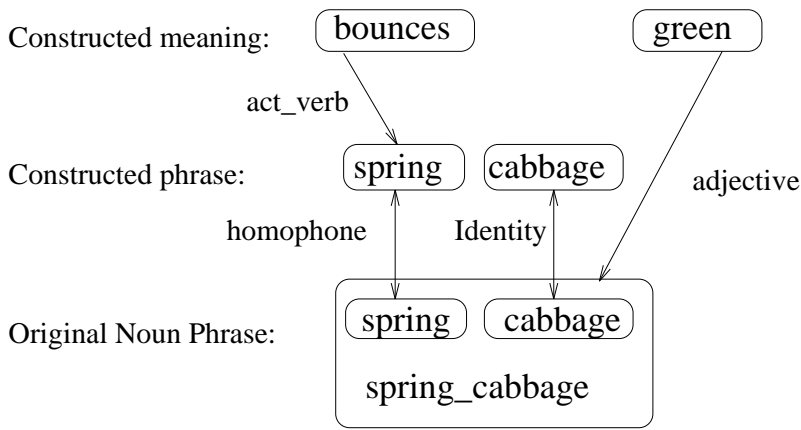}
\caption{A completely instantiated {\em lotus} schema}
\label{filled}
\end{figure}

The completely instantiated {\em lotus} schema in figure~\ref{filled}
could (with an appropriate template --- see subsection~\ref{tem}) be
used to construct the joke:

\begin{quote}
What's green and bounces? {\em A spring cabbage.} \cite{CAJB}
\end{quote}

\subsection{Templates}
\label{tem}

A template is used to produce the surface form of a joke from the
lexemes and relationships specified in an instantiated schema.
Templates are not inherently humour-related. Given a (real or nonsense)
noun phrase, and a meaning for that noun phrase (genuine or
constructed), a template builds a suitable question-answer pair. For
example, given the noun phrase ``black cat'' and a suitable
definition,\footnote{ Culture-specific of course --- to some people,
black cats are {\em lucky}.} some template might generate:

\begin{quote}
What kind of feline brings bad luck? {\em A black cat.}

\end{quote}

Whether the noun phrase is real or not, it is important that the
question provide the right amount of information --- too little would
fail to describe the `answer' completely, and too much would be
redundant. For example, the question part of:

\begin{quote}

What kind of feline drinks milk? {\em A black cat.}

\end{quote}
does not suggest in any way that the cat is black; whereas the question
part of:

\begin{quote}

What kind of unlucky dark feline animal brings bad luck and drinks milk?
{\em A black cat.}

\end{quote}
says too much.

Because of the need to provide an appropriate amount of information in
the riddle question, the choice of a schema constrains the choice of a
template somewhat. Some schemata construct `obscure definitions' (i.e.
definitions that do not strongly evoke the words in the constructed noun
phrase), and so require templates that provide more information in the
questions; whereas other schemata construct very clear definitions, and
so require templates that do not `give the joke away'. For example,consider
the joke:

\begin{quote}

What man claps at Christmas? {\em Santaplause.} \cite{CAJB}

\end{quote}
Using a template that provides too little information, this joke might
become:
\begin{quote}

What man claps? {\em Santaplause.}

\end{quote}
and using a template that provides too much information, it could be:
\begin{quote}

What bearded man in a red suit applauds at Christmas? {\em Santaplause.}

\end{quote}
\label{links-to-templates}
For this reason, every schema  has to be associated with a set of appropriate
templates.  This association is, strictly speaking, between templates and an
{\em instantiated} schema, not between templates and the skeletal
schema which defines the possible patterns of lexemes. This is because
the precise choice of relations for the
relations (e.g.\ ``characteristic'' links) will affect the appropriateness
of a template.
Conversely, one could say that the choice of template influences
the choice of particular relation for the links (this
is in fact how it is implemented  --- see Section~\ref{links}).
To be more precise, part of the knowledge which defines possible
riddles within our theory is a mapping which associates each
uninstantiated schema and selection of values for its underspecified
links with a set of templates.

Templates also state where a slot (variable) in a schema is to be
bound to a {\em sequence} of lexemes.  The required lexemes are indicated
by giving the lexical relations that they have to the lexeme at
the other end of the characteristic link.

Although templates are not humour-oriented by nature, it is clear that
riddles often use very regular forms, possibly because
such templates automatically provide an appropriate amount of information.
For example, questions of the form ``What do you get when you cross
\_\_\_ with \_\_\_ ?'' are quite common:

\begin{quote}

What do you get when you cross a sheep and a kangaroo? {\em A woolly
jumper.}

\end{quote}
In our implementation, templates embody such standard forms,
but this idea of standard forms is not essential to the concept of
templates. It is possible to imagine a template that generates punning
riddle surface forms `from scratch', as it were. Templates are simply a
mechanism for generating a suitable surface form for a
joke from the information in an instantiated schema.

\subsection{Conclusion}

In this section, we have developed a simple model for question-answer
punning riddles which use substitution as their main mechanism.  A
punning riddle is viewed as the surface manifestation of an
instantiated schema and an associated template (with its variable
portions filled from the schema). A schema defines the constraints on
lexemes which will produce possible riddles.

\section{The JAPE-1 computer program}
\label{program}
\subsection{Introduction}

We have implemented the model described in Section~\ref{theory}
in a computer program called \jape\ (Joke Analysis\footnote{``Analysis''
in the sense that it is an implementation of a model --- \jape\ does not
actively parse or analyse previously made jokes in any way.} and
Production Engine).
\jape\ is significantly different from other attempts to computationally
generate humour in three ways: its lexicon is humour-independent (i.e.
the structures that generate the riddles are distinct from the semantic and
syntactic data they manipulate); it generates riddles that are similar
on a strategic and structural level, rather than merely in surface form; and,
finally, \jape\ is an implementation of a model of riddles, so its success
or failure is of theoretical, as well as practical, interest.

Although our model could be applied to punning riddles in general, we
chose to concentrate on a subgroup. These riddles all:

\begin{itemize}

\item use word--word substitution as their main mechanism;

\item substitute phonologically identical (as opposed to phonologically
similar) words;

\item substitute into a noun phrase; and,

\item use the phrase thus constructed in the punchline (as opposed to
the question) part of the riddle.

\end{itemize}

These restrictions are chosen largely to reduce the number of schemata
and templates required, so that the important factors in
\jape's performance do not get lost in the noise. Also, the easy
availability of homophone lists makes word--word substitution schemata
easier to implement than those using word--syllable substitution or
metathesis (both of which use sub--word divisions). Although these
restrictions may seem excessive, they still describe approximately 100 of
the jokes in the Crack-a-Joke Book \cite{CAJB}.

\subsubsection*{Worked example}

In this section, as an illustrative example, we work through how \jape\
reproduces one of the Crack-a-Joke Book riddles, namely:

\begin{quote}
What do you get when you cross a sheep and a kangaroo? {\em A woolly
jumper.}
\end{quote}

\subsection{Top Level}
\label{top}

\jape's main mechanism attempts to construct a punning riddle based on a
common noun phrase. It has several distinct knowledge bases with which
to accomplish this task:

\begin{itemize}

\item a set of six schemata (see subsection~\ref{schemata});

\item a set of fourteen templates (see subsection~\ref{templates});

\item a lexicon, which contains humour--independent semantic and
syntactic information about the words and noun phrases entered in it
(see subsection~\ref{lexicon});

\item a homophone base, which contains pairs of phonologically identical
lexemes (see subsection~\ref{homonyms}).

\item a post--production checker, which applies some simple heuristics to
sift out some of the more obvious non--jokes (see subsection~\ref{checker})
\end{itemize}

As noted in Section~\ref{theory}
above, the homophone base is abstractly part of the lexicon, since it merely
embodies one more relationship between lexemes, but it was
computationally convenient to separate it out.

At the top level, \jape\ chooses an appropriate schema and an
associated template (as noted in Section~\ref{links-to-templates},
the choice of template is constrained by the choice of
schema, and the choice of template in turn specifies the content
of the underspecified links in the schema). It
instantiates the schema using information from the lexicon and the
homophone base, and fills the template, again using information from
the lexicon.  Filling a template produces a riddle in near--surface
form, which is tested by the post--production checker.

It is not inherent in \jape's design that these stages of joke
production happen in any particular order, although some choices do
constrain others.  In fact, in this implementation, the noun phrase is
chosen first (see subsection~\ref{algorithm} for details of the
algorithm).  Also, \jape\ optionally allows the noun phrase, the
schema, and/or the template to be specified by the user, so that the
program may be given as much, or as little, guidance as wished.

\subsection{Lexicon}
\label{lexicon}

The lexicon contains the lexemes used in constructing the jokes, and
syntactic and semantic information about them.

A {\em lexeme} is a symbol unique to one semantic interpretation of a
word. Each entry in the lexicon has one, and only one, corresponding
lexeme, which in turn is associated with a near--surface form. The
near--surface form of a word can be associated with several different
lexemes, and thus several lexicon entries.

A {\em near--surface form} is a piece of text (a word, phrase, sentence,
or complete riddle) in grammatical, understandable English; however, it
is not `pretty printed' (e.g.\ it may not have capitals at the beginning
of sentences, etc). In this implementation, a near--surface form is a
list of lower-case words. This is for programming convenience only ---
once a riddle has been generated and checked, it is changed into surface
form by the program.

The purpose of the lexicon is to store general ({\em not}
humour--oriented) syntactic and semantic information. Although this
lexicon was constructed specifically for \jape, the information
contained in it is general and neutral --- the joke--generating power
lies elsewhere in the program, particularly in the schemata and
templates, and the ties between them.

Although the lexicon stores syntactic information, the amount of syntax
used by the rest of the program is minimal. Because the templates are
based on common riddle forms, such as ``What do you get when you cross
\_\_\_ with \_\_\_?'', whole sentences need not be constructed `from
scratch'. For this reason, the only necessary syntactic information has
to do with the syntactic category, verb person, and determiner
agreement. Also, the lexicon need only contain entries for nouns, verbs,
adjectives, and common noun phrases --- other types of word
(conjunctions, determiners, etc) are built-in to the templates.
Moreover, because the model implemented in \jape\ is restricted to
covering riddles with noun phrase punchlines, the schemata require
{\em semantic} information only for nouns and adjectives  (see
Section~\ref{schemata} for more details).

Each lexeme can be considered to be a node in a network, linked to other
lexemes in the network via the semantic slots in its lexical entry. The
values in these semantic slots should be other lexemes with entries in
the lexicon. Syntactic slots, on the other hand, contain syntactic
information, not lexemes (see table~\ref{slots} for more details
on the available syntactic and semantic slots).

Although the values in the semantic slots {\em should} be other lexemes,
in some cases ({\sc has}, {\sc act\_obj}, {\sc location}, and {\sc
used\_to\_obj}) a semantic slot takes a near--surface form instead. This
is because we are interested in \jape\ as an implementation of a model
of riddles, rather than as a generator of syntactically--complex
sentences. In order to avoid complex (but uninteresting) syntactic
generation, the values in some semantic slots are chunks of text (i.e.
words put together grammatically in near--surface form) instead of
lexemes, so that they can be put directly into a template without
further syntactic manipulation. For example, the entry for the lexeme
{\bf lion} has, as its {\sc location} slot value,  the near--surface
form of ``in the jungle'', rather than the lexeme {\bf jungle}.

\begin{table}
\centering
\begin{tabular}{|l|l|l|} \hline\hline
{\em Slot} & {\em Used With} & {\em Allowed Values} \\ \hline\hline
{\em Syntactic Slots} \\ \hline\hline
{\sc category} & all entries & np, noun, adj, verb \\ \hline
{\sc written\_form} & all entries &
The near-surface form of the lexeme. For nouns, \\
& & this is taken by convention to be the singular \\
& & form, and for verbs, the infinitive. \\ \hline
{\sc vowel\_start} & np, noun, adj &
yes or no (does the near-surface form of the \\
& & lexeme start with a vowel?) \\ \hline
{\sc second} & verb &
The near-surface, second-person form of the verb. \\ \hline
{\sc third} & verb &
The near-surface, third-person form of the verb. \\ \hline
{\sc comp\_lex} & np &
A list of the lexemes that make up the noun \\
& & phrase. \\ \hline
{\sc countable}       & np, noun      &
yes or no (is the noun or np countable?) \\ \hline\hline
{\em Semantic Slots} \\ \hline\hline
{\sc class} & np, noun &
The immediate superclass of the lexeme (e.g.\ \\
& & for {\bf lemon}, {\bf fruit}) \\ \hline
{\sc spec\_is} & np, noun &
A lexeme that, when used to qualify the class \\
& & lexeme, defines the entered lexeme reasonably \\
& & precisely. (e.g.\ for {\bf lemon}, {\bf citrus})\\ \hline
{\sc is} & np, noun &
A lexeme that typically describes the entered \\
& & lexeme (e.g.\ for {\bf lemon}, {\bf sour}). \\ \hline
{\sc has} & np, noun &
Part(s) of the thing to which the entered lexeme \\
& & refers, in near-surface form. Should fill \\
& & ``It has \_\_\_.''. (e.g.\ for {\bf lemon}, ``pips''). \\ \hline
{\sc act\_verb} & np, noun &
A verb lexeme. Something the thing typically \\
& & does. (e.g.\ for {\bf chef}, {\bf cook})\\ \hline
{\sc act\_obj} & np, noun &
The near-surface form of the object of the \\
& & {\sc act\_verb} value. (e.g.\ for {\bf chef}, ``food'')\\ \hline
{\sc inact\_verb} & np, noun &
A verb lexeme. Something you typically do to \\
& & the thing. (e.g.\ for {\bf horse}, {\bf ride}). \\ \hline
{\sc location} & np, noun &
The near-surface form of its typical location. \\
& & (e.g.\ for {\bf horse}, ``in a pasture'')\\ \hline
{\sc used\_to} & np, noun &
A verb lexeme. Something the thing is typically \\
& &  used to do. (e.g.\ for {\bf spatula}, {\bf flip})\\ \hline
{\sc used\_to\_obj} & np, noun &
The near-surface form of the object of the \\
& & {\sc used\_to} value. (e.g.\ for {\bf spatula}, \\
& & ``pancakes'')\\ \hline
{\sc synonym} & np, noun, adj &
A lexeme of the same category as the entered \\
& & lexeme, which has a very similar entry --- \\
& & in particular, a lexeme's synonym's synonym \\
& & is itself. (e.g.\ for {\bf pillow}, {\bf cushion})\\ \hline
{\sc describes\_all} & noun, adj &
A lexeme which refers to a thing or class of \\
& & things which can (almost) always be described \\
& & by the entered lexeme. \\
& & (e.g.\ for {\bf slimy}, {\bf worm})\\ \hline
\end{tabular}
\caption{\label{slots} Lexicon Slots}
\end{table}

\subsubsection*{Worked example}
\label{we:entries}

The phrase ``woolly jumper'' has the following entry, where
bold-face entries indicate lexemes:
\begin{quote}

lexeme = {\bf woolly\_jumper} \\
category = np \\
written\_form = ``woolly jumper'' \\
comp\_lex = {\bf woolly}, {\bf jumper\_1} \\
vowel\_start = no \\
countable = yes \\
class = {\bf sweater} \\
inact\_verb = {\bf wear} \\

\end{quote}
Someone else defining the word could have provided slightly different
semantic slot values (e.g.\ a British word--definer  would  probably not
have used the word ``sweater'') or filled in more slots. However, any
slot values which fit the descriptions in table~\ref{slots} should
produce recognizable jokes.

``Woolly'', an adjective, has the following entry:
\begin{quote}

lexeme = {\bf woolly} \\
category = adj \\
written\_form = ``woolly'' \\
vowel\_start = no \\
describes\_all = {\bf sheep} \\
synonym = {\bf fuzzy}

\end{quote}
Again, another definer could make a slightly different entry, which
would produce a different end joke.

Finally, the word ``jumper'' has two entries, one for each
meaning:
\begin{quote}

lexeme = {\bf jumper\_1} \\
category = noun \\
written\_form = ``jumper'' \\
vowel\_start = no \\
countable = yes \\
class = {\bf clothing} \\
spec\_is = {\bf warm} \\
synonym = {\bf sweater} \\

\end{quote}
\begin{quote}

lexeme = {\bf jumper\_2} \\
category = noun \\
written\_form = ``jumper'' \\
vowel\_start = no \\
countable = yes \\
describes\_all = {\bf kangaroo} \\
act\_verb = {\bf leap}\\

\end{quote}

Each of the lexeme values in the semantic slots of the above entries has
an entry of its own.

\subsection{Homophone Base}
\label{homonyms}

The homophone base is simply a list of homonym (different spelling)
pairs and alternate--meaning (same spelling) pairs. These lexemes are
available to be used as the ends of a {\em homophone} link, when
instantiating a schema (see section~\ref{schemata}).

Homonyms are words that are {\em phonologically} identical, but have
different spellings.  Homonyms should also have different meanings;
that is, spelling variants, such as ``humor'' and ``humour'', are not
considered to be homonyms.  An alternate--meaning pair, on the other
hand, is a pair of lexemes that have identical near--surface forms,
but different semantic entries (sometimes known as {\em homographs}).
For example, the lexeme {\tt sole\_1}, which refers to a kind of fish,
and {\tt sole\_2}, an adjective synonymous with ``only'', are
alternate meanings; however, {\tt shower\_1}, a light rain, and {\tt
shower\_2}, a bathroom device, are not, since they both refer to water
falling from above. In \jape\, the entries for the two lexemes must be
{\em completely} different for them to be an alternate--meaning pair.

\jape's homophones are from a list \cite{Townsend} of homophones in
American English, which has been shortened considerably for our
purposes. Removed from the list were:

\begin{itemize}
\item pairs including a proper noun (e.g.\ ``Cain'' and ``cane'')
\item pairs including an obscure word (e.g.\ ``buccal'' and ``buckle'')
\item pairs which are not homonyms in British English (e.g.\ ``balm'' and
``bomb'')
\item pairs including words which are neither adjectives nor nouns
\item pairs including  abstract nouns
\item pairs whose meanings are often confused (e.g.\ ``acclamation'' and
``acclimation'')

\end{itemize}

This version of \jape\ treats homonym pairs and alternate-meaning pairs as
equivalent, since they seem to play the same role in the type of riddle
we are trying to reproduce. However, one can imagine types of joke in
which they are not equivalent, so they have been kept distinct in the
homophone base.
We shall take the liberty of referring to both homonyms and alternate
meanings as ``homophones'' from now on.

\subsubsection*{Worked example}

The two lexemes {\bf jumper\_1} and {\bf jumper\_2} are entered in the
homophone base as an alternate--meaning pair, since the semantic parts of
their entries (see subsection~\ref{we:entries}) are completely
different.

\subsection{Schemata}
\label{schemata}

\begin{figure}[htb]
\epsfxsize = 0.75\textwidth
\epsfbox{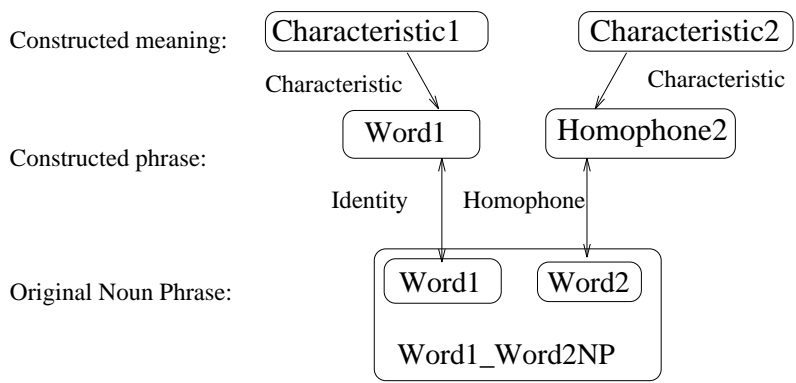}
\caption{The uninstantiated {\em jumper} schema}
\label{jumper}
\end{figure}

\begin{figure}[htb]
\epsfxsize = 0.75\textwidth
\epsfbox{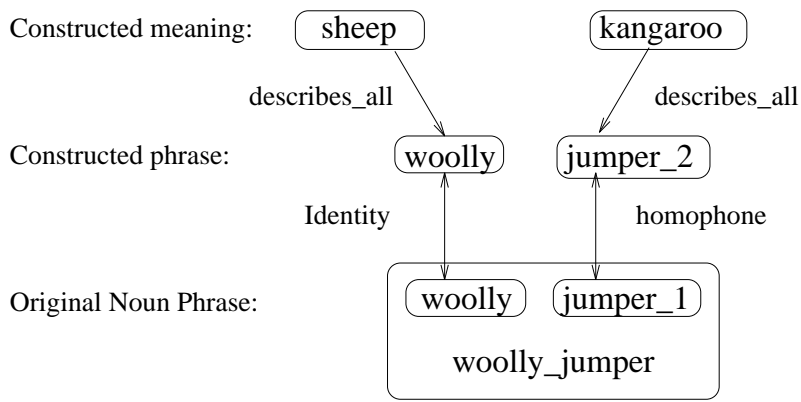}
\caption{The instantiated {\em jumper} schema, using the noun phrase
``woolly jumper'' and the {\em syn\_syn} template. Used to generate the
joke: What do you get when you cross a sheep and a kangaroo? {\em A
woolly jumper}.}
\label{jumpfill}
\end{figure}

The {\em jumper} schema\footnote{Although it is not accidental that this
schema's name is one of the words in the punchline of the worked
example, it is just a mnemonic. The {\em jumper} schema can, of course,
generate other jokes, as we later demonstrate.} shown in
figure~\ref{jumper} is uninstantiated, in that the actual lexemes to be
used to construct the joke have not yet been specified, and some of the
relationships (the characteristic links) between the lexemes are still
very general. The schema in figure~\ref{jumpfill} is the same schema,
completely instantiated.

\begin{figure}[htb]
\epsfxsize = 0.75\textwidth
\epsfbox{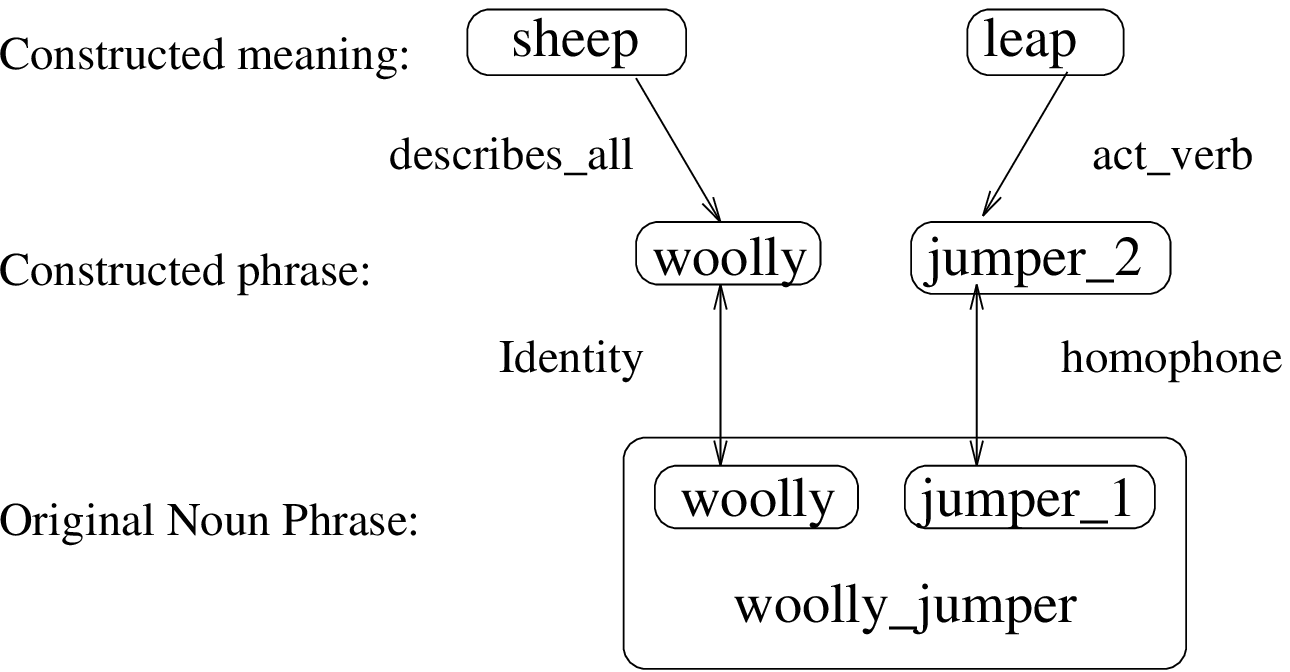}
\caption{The instantiated {\em jumper} schema, using the noun phrase
``woolly jumper'' and the {\em syn\_verb} template. Used to generate the
joke: What do you call a sheep that can leap? {\em A woolly jumper}.}
\label{jumpfill2}
\end{figure}

\label{links}

These three links (characteristic, identity, homophone) are all that is
required to specify a wide range of riddle schemata. However, if
\jape\ were to be expanded to handle other kinds of joke, more links
would have to be added. For example, if \jape\ had to produce
spoonerisms, a {\em rhymes} link (at least) would be necessary.

Characteristic links indicate that the lexemes chosen for the schema
must be linked via some relation based on the lexical entries.
The simplest values for characteristic links, therefore, are the
actual slot labels used in the lexicon:

\begin{tabular}{ll}
$\bullet$ inact\_verb & $\bullet$ location\\
$\bullet$ synonym & $\bullet$ describes\_all\\
$\bullet$ class & $\bullet$ spec\_is\\
$\bullet$ is & $\bullet$ has\\
$\bullet$ used\_to & $\bullet$ used\_to\_obj\\
$\bullet$ act\_verb & $\bullet$ act\_obj\\
\end{tabular}

Characteristic links can also join a lexeme to a {\em concatenation} of
lexemes. For example, one template uses a sequence of two semantic slot values,
namely the {\em spec\_is} and {\em class} entries.
This link would join the lexeme {\bf jumper\_1} to the two lexemes {\bf
warm} and {\bf clothing}. The information about what lexemes to concatenate
is held in the template, and is tied to the exact choice of relation
made for the characteristic link.

We have found it useful, in describing \jape's mechanisms,
to make a distinction between  two sorts of lexemes within
a schema. There are {\em key} lexemes, which can be instantiated
early on in the process because either they are part of the
genuine noun phrase, or they are linked to such lexemes by
well-specified links such as {\em Identity}.
There are also {\em characteristic} lexemes, which are linked
only by {\em Characteristic } links, and so cannot be chosen until
the template is chosen.

The characteristic links in a schema are instantiated as soon as a
template is chosen, and any lexemes still uninstantiated in the schema
(the characteristic lexemes) are inserted as the template is filled.
However, it is important to distinguish between the pieces of text
generated by the template, and the lexemes in an instantiated schema,
because it is the lexemes which are checked for suitability later (see
section~\ref{checker}).

\subsubsection*{Worked example}

The example we are working through uses the {\em jumper} schema, as
shown in figure~\ref{jumper}.  In figure~\ref{jumpfill}, an
instantiated {\em jumper} schema, the key lexemes have been slotted in
appropriately. Also, both of the {\em characteristic} links have been
instantiated to {\em describes\_all}, and the characteristic lexemes
have been instantiated to {\bf sheep} and {\bf kangaroo}, from the
lexical entries for {\bf woolly} and {\bf jumper}.

It is important to understand that this particular instantiation of
the {\em jumper} schema is a result of two factors: the chosen noun
phrase ({\bf woolly\_jumper}), and the chosen template (in this case,
{\em syn\_syn}, described further in subsection~\ref{templates}). The
choice of template determines how the characteristic links are
specified, and thus the values for the characteristic lexemes.
For example, if a different template, {\em syn\_verb} (described in
subsection~\ref{templates}), were chosen, the {\em jumper} schema
would be instantiated in a slightly different way --- one of the
characteristic links would be instantiated to {\em act\_verb} instead
of {\em describes\_all}, and thus that characteristic lexeme would be
{\bf leap} instead of {\bf kangaroo} (see figure~\ref{jumpfill2}).

\subsection{Templates}
\label{templates}

As described in section~\ref{theory}, a template constructs a suitable
question--answer pair from the information contained in a schema. The
question uses the `meaning' of some nonsense phrase to refer to that
nonsense phrase in the answer, or punchline. Both the nonsense phrase
and its `meaning' are built by the instantiation of a schema.
Riddles tend to use one of a limited number of common forms; for
example, questions of the form ``What do you get when you cross \_\_\_
with \_\_\_?'' come up quite often. \jape\ exploits these
common forms, in order to cut down on the syntactic information required
to generate a grammatical surface-form riddle.
\jape's templates have two important parts:
\begin{itemize}

\item specifications for the underspecified links in the schema, and

\item near--surface form question--answer pairs containing blanks, which
are to be filled by near--surface form text fragments. These text
fragments are generated by the template, from the lexemes in the
instantiated schema.

\end{itemize}
(Also, where a particular choice of relation involves instantiating
a variable to a  particular {\em sequence} of lexemes, that will be indicated).
For example, the {\em syn\_syn} template can be stated informally as:
\begin{quote}

Use relations from the set \{ {\bf spec\_is\_class, describes\_all, synonym}\}
to instantiate the characteristic links; for {\bf spec\_is\_class},
form a sequence from the two lexemes that are linked to the key lexeme by
{\bf spec\_is} and by {\bf class}.
 \\
 \\
What do you get when you cross {\bf [text fragment generated from the first
characteristic lexeme(s)]} with {\bf [text fragment generated from the second
characteristic lexeme(s)]}? \\
{\em [the near--surface form of the constructed noun phrase].}

\end{quote}

and the {\em syn\_verb} template  can be summarised:

\begin{quote}
Instantiate first characteristic link to an element of
\{ {\bf spec\_is\_class, describes\_all, synonym}\},for {\bf spec\_is\_class},
form a sequence from the two lexemes that are linked to the key lexeme by
{\bf spec\_is} and by {\bf class}; instantiate second characteristic
link to {\bf act\_verb}\\
 \\
What do you call {\bf [text fragment generated from the first characteristic
lexeme(s)]} that {\bf [text fragment generated from the second characteristic
lexeme(s)]}? \\
{\em [the near--surface form of the constructed noun phrase].}

\end{quote}

In this implementation, the text fragments are generated using a
simple Prolog definite-clause grammar \cite{bratko}, which simply puts
the lexemes together in a syntactically correct near--surface form.

\subsubsection*{Worked example}

If the {\em syn\_syn} template is chosen, then the {\em jumper} schema
is instantiated to the schema in figure~\ref{jumpfill}. Then, the
information in that instantiated schema is given to the {\em syn\_syn}
template, described above, which generates the near--surface form of:

\begin{quote}

What do you get when you cross a sheep and a kangaroo? {\em A woolly
jumper.}

\end{quote}

Similarly, if {\em syn\_verb} is the chosen template, the {\em jumper}
schema is instantiated to the schema in figure~\ref{jumpfill2}. The
information in that instantiated schema is then used by the {\em
syn\_verb} template to generate the near--surface form of:

\begin{quote}

What do you call a sheep that can leap? {\em A woolly jumper.}

\end{quote}

\subsection{Post-production checking}
\label{checker}

Although \jape\ generates pieces of text that are recognizably jokes,
they are not all {\em good} jokes (see Section~\ref{evaluation} for an
evaluation of \jape's performance). It is clear that some kind of
post-production heuristic checking is needed, to put the jokes in order
of quality, and perhaps remove those at the bottom of the pile.
At present, only two such checks have been implemented. The first is
that none of the lexemes used to build the question and punchline are
accidentally identical; that is, \jape\ checks that lexemes in an
instantiated schema are identical only if linked by an identity link.
This is to prevent `jokes' like the following:

\begin{quote}

What do you get when you cross a sheep and a jumper? {\em A woolly
jumper.}

\end{quote}
Such a `joke' might arise from loops in the lexicon --- entries
referring to each other in an unexpected way.

The second post-production check is that the lexemes used to build the
nonsense noun phrase (i.e.\ the punchline) do not build a genuine common
noun phrase. This is to prevent \jape\ from accidentally coming up with
{\em sensible} question--answer pairs, such as:

\begin{quote}

What do you call a cylindrical drink container? {\em A coke can.}

\end{quote}
It is difficult to imagine why \jape\ would produce such a piece of text;
however, it is clearly not a joke, and should be sifted out.

\subsection{JAPE's basic algorithm}
\label{algorithm}

In this design, there are several points at which a choice has to be
made, a schema instantiated, or a template filled. \jape\ tackles these
tasks in the following order:

\begin{enumerate}
\item Choose a common noun phrase from the lexicon.
\item Choose an appropriate schema, into which the phrase can fit.
\item Fit the phrase's constituent lexemes into the schema.
\item Instantiate the key lexemes in the schema.
\item Choose an appropriate template, thus instantiating the
characteristic links.
\item Generate the near--surface form of the riddle, instantiating
characteristic lexemes in the process.
\item Check that the generated joke is neither repetitive, nor a
sensible question-answer pair.
\end{enumerate}
However, this exact ordering is a feature of the implementation, rather
than the underlying model.

\section{Factors likely to influence JAPE-1's performance}
\label{hypotheses}

In \jape's design as it stands, there are places open for the
introduction of heuristics (rules of thumb) for improving joke
quality, both during joke production and the post-production sifting of
\jape's output. No such heuristics were introduced before testing, so
that they would arise from an evaluation of \jape's unfiltered output,
rather than pre-run guesswork. Nonetheless, it was possible, prior to
the evaluation, to make
certain predictions about \jape's performance which, if
confirmed, could give rise to heuristics in later versions of the
program. Some of the suggested heuristics would {\em order} the output
of the program; others would {\em trim} the output, by reducing the
number of jokes produced.

All of these conjectures were made after the lexical data was gathered,
but before \jape\ was run to produce the set of jokes to be judged (see
section~\ref{methodology} for details of the methodology).

\subsection{The Lexicon and the Homophone List}

It was anticipated that the quality and content of the lexicon
would largely determine the quality of the jokes produced.

Each of the steps in \jape's joke production rely on the lexicon being
clear, precise, accurate, and compatible with \jape's templates and
schemata. For this reason, it was anticipated that the following
requirements would be important:

\begin{itemize}

\item Semantic information should be included in the lexicon only if it
is typical of the word being entered. For example, dogs do sleep, but if
the entry for ``dog'' includes that information, then \jape\ is likely to
ask questions like ``What sleeps and \ldots'', expecting ``dog'' to be
among the concepts evoked in the mind of the reader.

\item The information in the lexicon should be {\em common knowledge}.
This applies both to the words chosen to be entered, and the entries
they are given. It would be no use \jape\ making a joke about a
horse's frog, if most people do not know that ``frog'' can refer to
part of a horse's foot, as well as to a green amphibian.

\item Jokes should use concrete words (e.g.\ ``wooden'', and ``cat''),
rather than abstract words (e.g.\ ``happiness'', ``attitude''). This is
because the constructed concept is more likely to be funny if it can be
visualized. For example, the idea of a ``toe truck'' probably has more
humour potential than that of an ``optical allusion''.

\item  Jokes should avoid using very general words (e.g.\ ``structure'',
``substance'', ``object'') because such words have a huge number of
possible instances. For example, although a Buddhist monk is indeed a
person, the word ``person,'' without any other information, is unlikely
to bring the image of a Buddhist monk to mind --- so a joke depending on
that evocation would probably fail.

\item Homophone pairs should not be easily confused (i.e.\ the distinctness
of their meanings should be common knowledge). For example, a pun based on the
homophones ``aural'' and ``oral'' would probably not work very well, as
these words are often misspelled and misused.

\end{itemize}

\subsection{Schemata}

Some schemata substitute only one homophone; others substitute
homophones for both the words in the phrase.
The more homophone links in the schema, the more information should be
available for the template to use in the question part of the riddle.
It was anticipated that any schema which did not provide enough
information for the riddlee to `solve' both homophone-substitutions,
would therefore produce incomprehensible jokes.

On the other hand, a schema which both uses two homophone-substitutions,
{\em and} provides the template with a great deal of information about
the ``meaning''
of the nonsense phrase (e.g.\ our {\em ginger} schema)
was expected to produce the best jokes.
However, it would also produce the
fewest jokes, for the same reason --- few common noun phrases have
constituent words which both have homophones, and few of our lexicon
entries contain enough semantic information to fill this schema.

It was suspected that schemata which rely on good, clear lexicon
entries for noun phrases would produce slightly worse riddles,
as such entries are difficult to make (because, in
general, noun phrases describe more complex concepts than individual
words).

If a schema may create confusion between the meaning of the head noun
of the punchline and the meaning of the whole punchline noun phrase
(e.g.\ the {\em elan} schema), this may give poor results.

It was anticipated that all the schemata would produce jokes, but for
the above reasons, they were expected to be successful in this
(descending) order:

\begin{itemize}

\item {\em ginger}
\item {\em woolly} and {\em jumper}
\item {\em lotus}
\item {\em elan}
\item {\em double\_pun}

\end{itemize}

\subsection{Templates}

The central part of the punchline, in all templates, is a noun phrase
consisting of a noun and a modifier.
Since the word order of the punchline does contain some semantic
information, namely, which of its words is the object and which word
describes that object, it is important for the question to reflect that
information. For example, if the punchline is to be ``lemon aide'', then
the question should describe a kind of assistant, rather than a kind of
citrus fruit; that is,

\begin{quote}
What do you call a sour assistant? {\em A lemon aide.}
\end{quote}
is a more coherent riddle than:

\begin{quote}
What do you call a helpful fruit? {\em A lemon aide.}
\end{quote}

The characteristic order in the question need not be the same as that in
the punchline, but it should describe a type of whatever the second word
is. For example, the {\em use\_syn} template (``What do you use to \_\_\_ a
\_\_\_?'') could be used in non-joking communication as follows:

\begin{quote}
What do you use to fight a fire? {\em A fire fighter.}
\end{quote}

A coherent joke should have a similar structure, for example:

\begin{quote}
What do you use to wash a harness? {\em A bridle shower.}
\end{quote}
rather than the reverse:

\begin{quote}
What do you use to steer drizzle? {\em A bridle shower.}
\end{quote}

For this reason, it was anticipated that the  templates
which conformed to this pattern would
be more successful than templates which use the
lexemes in the opposite order.

We expected that another important factor in the success of a template
would be the amount of information a question generated by the template
would provide about the punchline. If it were to provide too much, the
punchline would be given away; if it were to provide too little, the joke
would be incomprehensible.

\subsection{Schema-Template Pairs}

It was anticipated that certain schemata would work well with some
templates, and not so well with others. It was difficult to predict
which would be the successful pairs; however, we expected the
distinction to be clear. In fact, we predicted that the elimination of
certain schema-template pairings would improve the data considerably.

\subsection{Post-Production Checking}
\label{pp-checking}

At the time
of testing, the only post-production checks in \jape\  were that
the punchline of the riddle was not a
real noun phrase (i.e.\ that it was not in the lexicon), and that none of
the key lexemes used to build the riddle were accidentally identical (i.e.
they were identical only if linked by an identity link). Other
less obvious checks were not included in pre-evaluation \jape\, in the
hope that the evaluation would reveal their heuristic power so that they
could be added later.

Some good heuristics were expected to be:

\begin{description}

\item[question length:] Questions that are too long seem unwieldy;
questions that are too short don't provide enough information to suggest
the punchline.

\item[alliteration and rhyming:] Punchlines made up of alliterative or
rhyming words are often more successful. Spoonerisms and
related types of joke rely largely on this effect for their humour.

\item[`funny letters':] It is a commonly-held belief among stand-up
comedians that certain letters (particularly ``k'', ``q'', ``v'', ``w'', and
``z'') are inherently funny --- of a pair of synonyms, the one containing
the most such letters is funnier. This may be a myth; on the other
hand,  it is may explain why one joke is a little bit funnier that
another similar one.

\item[subject matter:] Jokes that have `funny' subjects (sex,
religion, politics, etc.) are generally funnier than similar jokes
about more mundane issues.

\item[accidental associations:] It was expected that some jokes produced
by \jape\ would be funny in unforeseen ways. For example, the question part
of the riddle could be syntactically ambiguous, thus making the
punchline a more humorous answer; or a semantic link not included in
the lexicon could be suggested by an accidental juxtapostion of words in
the riddle.

\end{description}

Obviously, some of these potential heuristics are easier to implement
than others. For example, if accidental associations were easy to spot,
they would not be accidental --- they would be included in \jape's lexicon,
schemata, or templates. Also, none of these post-production heuristics
could absolutely determine whether or not a given joke is funny;
however, they could be used to partially order \jape's output in terms
of quality.

\section{The evaluation procedure}
\label{methodology}
\label{evaluation}

The purpose of the evaluation was twofold. Primarily, it was to point the
way to improvements in the theory behind the project. This information
could then be used both to improve \jape's design, and to suggest
directions for further research. This evaluation also provided
both a relatively unbiased input (in the form of lexicon entries
provided by volunteers) for \jape\ as it stands, and a rough assessment
of \jape's current abilities.

This evaluation is {\em not} intended to be a rigorous examination of
the `humour value' of \jape's output, for several reasons. Firstly, it
would be a project in and of itself to design a rigorous experiment to
test the humour content of text jokes. Secondly, \jape\ should be seen
as an initial exploration, rather than a final product. Although a
statistically correct test would be valuable, it would require a huge
amount of data and a large number of volunteers --- effort that would
be wasted when some of the more obvious flaws in \jape's design were
fixed.

There are three stages to this evaluation: {\em data acquisition}, {\em
common knowledge judging} and {\em joke judging}. During the data
acquisition stage, volunteers unfamiliar with \jape\ were asked to make
lexical entries for a set of words given to them. These definitions were
then  sifted by a `common knowledge judge', and entered into \jape's
lexicon. The jokes produced by \jape\ from these sets of words were then
judged by a different group of volunteers. Their opinions (both quantitative
and qualitative) were then analysed, and compared to the hypotheses made
in section~\ref{hypotheses}.

\subsection{Lexical Data Acquisition Stage}

This is the phase in which the words \jape\ uses to build jokes were
defined. The lexicon is intended to be neutral, with all \jape's
joke-making knowledge stored in the schemata and templates. It is
therefore important that unbiased volunteers define words for \jape\, as
it is entirely possible that someone familiar with \jape\ would
(unconsciously) bias their entries towards making jokes.

Before the volunteers could define the words, however, an appropriate set
of words had to be chosen. Allowing the volunteers to randomly define any
words they like might well be less biased; however, they would have to
work their way through most of the English language before they stumbled
upon enough words that \jape\ could use.

First, the homophones to be defined were picked from a list of homophones,
supplied by \cite{Townsend}. Homophone pairs were eliminated from this
list if they were:

\begin{itemize}
\item not adjectives or nouns
\item abstract or obscure
\item often confused with each other
\item alternate spellings of a word
\item differed mainly in their syntactic category (e.g.\ ``bare'' the
adjective and ``bare'' the verb)
\end{itemize}

Common noun phrases which contain at least one word on the homophone list
were then added to the list of lexemes to be defined. Finally, the other
words used in the common noun phrases were added to the list. Twenty-one
noun phrases and fifty-nine words were then on the list, adding up to
eighty lexemes needing to be defined.

The list was divided into ten sets of eight lexemes each, with no two
words from the same noun phrase or homophone pair in each set. Ten
volunteers were each given a set of lexemes and instructions on how to
define them in accordance with the specification of \jape's lexicon
(see Table \ref{slots} earlier). In their
instructions, it was emphasized that they should provide only {\em
typical}, {\em specific} information about the lexemes they were to
define.

\subsection{Common Knowledge Judging}

After the lexical data had been collected,
it was discovered that some of the volunteers had not
followed the instructions, and had tried to fill every available slot in
each entry. Moreover, some volunteers had left words undefined, while
others were excessively creative in their use of English.

As a lexicon containing these entries would not meet \jape's
specifications, a `common knowledge judge' was recruited to sift the
entries. She could take only the following actions:
\begin{itemize}
\item {\bf veto} slot values

\item {\bf veto} an entire entry

\item {\bf move} a slot value into a different slot

\item {\bf define} a lexeme, only if it had no entry, either because the
original definer did not know what the word meant or because the
common-knowledge judge had vetoed the entry. In this case, the
experimenter (KB) would then have veto
power only over the new entry. (This did not happen often. Two lexemes
were left undefined by the original volunteers, and the common knowledge
judge vetoed only two more complete entries.)

\end{itemize}

After the entries had been sifted in this way, they were entered into
\jape's lexicon. \jape\ then produced a set of 188 jokes in near-surface form.

\subsection{Joke Judging Stage}

The 188 jokes produced by \jape\ were put into surface form and
distributed to fourteen `joke judges' recruited from
the experimenter's acquaintances. As each joke was given to two
judges, each volunteer had about 25 jokes to judge.

The questionnaire given to the judges had three sections. The first
was a list of jokes, each based on a different noun phrase, for them to
rate from 0 to 5 on the following scale:

\begin{enumerate}
\setcounter{enumi}{-1}
\item Not a joke. Doesn't make any sense.
\item A joke, but a pathetic one.
\item Not so bad.
\item OK. Might actually tell it to someone.
\item Quite good. Might tell it to someone and not get hit.
\item Really good.
\end{enumerate}

The second was a list of several sets of jokes for them to rate from
zero to five, {\em and} put in order within each set. All the jokes in a
set were based on the same noun phrase. This section was necessary
because some noun phrases produced a huge number of jokes, while others
produced very few, making an even distribution impossible.
Unfortunately, this high concentration of similar jokes made it very
likely that the judges would suffer from `joke fatigue', and dislike
the repetition.

The third section asked for qualitative information, such as how the
jokes might be improved, and if they'd heard any of the jokes before.

Each volunteer completed their questionnaire, and the results were
collected, collated and analysed.

\subsection{Problems with Methodology}

As mentioned earlier, this testing is not meant to be statistically
rigorous. However, when it comes to analyzing the data, this lack of
rigour causes some problems.

A statistically sound analysis requires a lot of data. A lot of data, in
this case, requires a lot of volunteers, and a lot of time to analyse
the data. During this project, neither was available. Unfortunately,
this dearth of data means that it is very difficult to draw conclusions
from the data we do have. For example, if a certain schema-template pairing
produces only one joke, due to the small size of the lexicon, it is
impossible to decide if that joke (and its score) is typical of that
pairing.

Another problem was the lack of control group. We suspect that jokes of this
genre are not very funny even when they are produced by humans; however,
we do not know how human-produced jokes would fare if judged in the same
way \jape's jokes were, so it is difficult to make the comparison. There
was no control group for two reasons: the number of volunteers (having
an adequate control group would double the number required), and the
difficulty of choosing which jokes would be in the control group.

The common knowledge judge was used to make up for the fact that there
were not enough volunteers to make multiple entries for each word, which
could then be compared. If there had been more volunteers, the
intersection of the entries for a particular word could then be taken as
the `common knowledge' about that word, and there would be no need for a
common knowledge judge.

The makeup of this group of volunteers was a bit strange, which could
well have affected the data. Due to the method of distribution of
questionnaires (e-mail to friends and associates), all the volunteers
were adults with regular access to a computer. Moreover, all the
volunteers were either comedians or involved in artificial intelligence
and so had expertise in one aspect or another of this project. This
undoubtedly influenced their judgement.

Ideally, with hindsight, the lexicon entries would have been made by a
large group of adults without any particular interest in computers or
comedy, and the intersection of these entries put into \jape's lexicon.
Alternatively, the lexical data could have been taken from some standard
lexicon. The resulting jokes would then have been mixed with similar
jokes (from \cite{CAJB}, for example), and then all the jokes would
have been judged by a group of schoolchildren, who would be less likely
to have heard the jokes before and more likely to appreciate them.

\section{Analysis of Results}
\label{results}

\subsection{Summary}

\input{epsf}
\begin{figure}[htb]
\mbox{\epsfxsize=4in
\epsffile{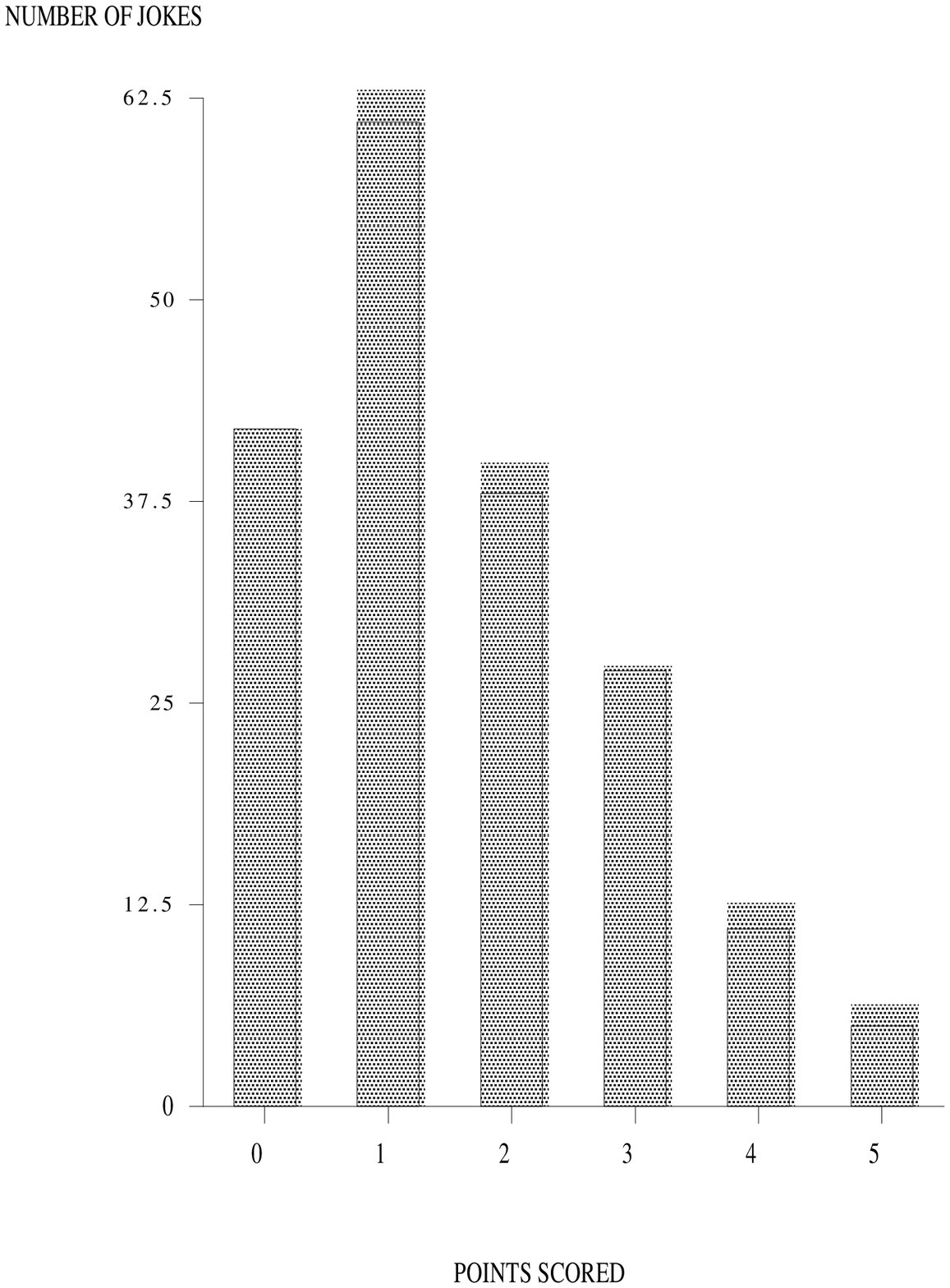}}
\caption{The point distribution over all the output}
\label{pointdis}
\end{figure}

The results of the testing are summarised in
figure~\ref{pointdis}, with fuller details in Appendix
\ref{full_result_summary}.
The average point score for all the jokes \jape\ produced from the lexical
data provided by volunteers is 1.5 points, over a total of 188 jokes.
Most of the jokes were given a score of 1. Interestingly, all of the nine jokes
that were given the maximum score of five by one judge, were given low
scores by the other judge --- three got zeroes, three got ones, and
three got twos.
The comments made on the questionnaire are discussed in
more detail below.

Overall, the current version of \jape\ produced, according to the scores
the judges gave, ``jokes, but pathetic ones''. On the other hand, the
hypotheses in section~\ref{hypotheses} were, for the most part, correct,
and the results show that implementing even the simplest of these
hypotheses would improve the resulting jokes significantly. Moreover,
the top end of the output (i.e.\ those jokes that would survive the
implementation of trimming heuristics) are definitely of Crack-a-Joke
book quality, including:

\begin{quote}
What do you call a murderer that has fibre? {\em A cereal killer.} \\
What kind of pig can you ignore at a party? {\em A wild bore.}\\
What kind of emotion has bits? {\em A love byte.}
\end{quote}

\subsection{The Lexicon and the Homophone List}

It is clear from the results (in particular, from the comments) that the
quality of the lexicon greatly influenced the quality of the resulting
jokes. Although we tried to ensure that the volunteer word definers
understood and followed the lexicon requirements, and attempted to sift out
inappropriate data, several `bad' entries (or parts thereof) survived to
be used in joke construction. It is difficult to trace exactly
what went wrong in a failed joke; however, the judge's comments suggest
that many of the flaws arose from entries in the lexicon that did not
meet the criteria described in section~\ref{hypotheses}:

\begin{itemize}

\item {\bf Semantic information should be included in the lexicon only if it
is typical of the word being entered.}
With reference to the rather poor joke:
\begin{quote}
What do you use to sniff a drilling tool? {\em A wild bore.}
\end{quote}
one judge asked, ``Does `to wild' mean `to sniff'?'' clearly trying to
understand what sniffing has to do with the punchline. It came from the
entry for ``boar'': the definer wanted to express the idea that boars
can be used to sniff for truffles. Unfortunately, sniffing as such is
not typical of boars, and so does not bring them to mind at all. The
common knowledge judge did not spot this particular entry.

\item {\bf The information in the lexicon should be {\em common knowledge}.}
This cropped up several times. Regarding these two jokes:
\begin{quote}
What do you use to clothe a coniferous tree? {\em A fir coat}.
\end{quote}
\begin{quote}
What do you call a passenger ship you can drink? {\em Ferry
liquid}.\footnote{``Fairy Liquid'' is a well-known brand of detergent in
Britain.}
\end{quote}
\noindent
one judge remarked:
\begin{quote}
\ldots the fir coat one is funny because most people would know
what a coniferous tree is \ldots whereas the ferry liquid
[joke] isn't very obvious.
\end{quote}
A related problem was that the judges were from all over the world, and
dialects differ.
A Canadian judge said that there were
\begin{quote}
\ldots too many `Britishisms' not common to North American English, eg
stag, hart, and queue.
\end{quote}
while a British judge pointed out an `Americanism':

\begin{quote} The bridal shower jokes rely on knowing that a wedding party is
called a bridal shower in North America.
\end{quote}

\item {\bf Jokes should use concrete words \ldots rather than abstract words.}

The only abstract words included in this lexicon were ``love'' and
``number'', neither of which caused any problems. In fact, the jokes
they were used in got above average scores.

The difficulty anticipated with abstract words was that, in general,
they do not evoke strong images the way more concrete words do.
Perhaps ``love'' and ``number'' are evocative enough to avoid this
problem.

\item {\bf Jokes should avoid using very general words.}

Most overly general words were sifted out of the lexicon by the
common-knowledge judge, as she was specifically instructed to watch
for them. For example, if she could think of a more immediate {\bf
class} value for the lexeme, she would veto the value given. Too general
words were not, therefore, a particular problem.

\end{itemize}

\subsection{Schemata}

As predicted, there was not a lot of difference in the average point
scores of the three best schemata, {\em lotus}, {\em woolly} and {\em
jumper}. For jokes using these schemata, the important factors affecting
their scores were the choice of template and/or noun phrase used to
build the joke.
The {\em elan} schema did not do as well as the others, as expected,
because there is often some confusion over
which words are referring to which concepts
This does not mean, however, that the {\em elan} schema should be
discarded, but rather that the templates with which to pair it must be
carefully chosen, so that the `meaning' of the pun is clearly expressed.

The schema which made two substitutions and gave little information
about the described concept, did poorly, as expected.  This suggests
that a joke based on {\em two} homophone substitutions into a noun
phrase needs more information to be provided in the question part of
the joke, if it is to be comprehensible.

The other double-substitution schema, which we predicted would produce
the fewest, but best, jokes, failed to produce any jokes at all. This is
due to the fact that  not only does the schema requires a noun phrase
into which {\em two} homophones can be substituted, but also its only
compatible template requires several
lexemes to be fully instantiated, and there was insufficient
lexical data to achieve this.

\subsection{Templates}

The templates varied a great deal in the quality of jokes they produced.

The two best templates were (informally)
``What do you get when you cross a ...
with a ...?'' and ``what do you call a ......?''.
This is, we suspect, because they provide the right amount of
information in the question --- not so much that the punchline is given
away, and not so little that the joke is incomprehensible.

Certain templates fared much better than similar templates that used the
lexemes provided by the schemata in the opposite order, as predicted,
apart from the ``what do you call .... that ....?'' template,
in which the order seemed not to be significant.

A general comment that was made several times was that some of the questions
were not `logically coherent' in some sense. This incoherence was often
the result of using one of the templates with bad word ordering. For
example, the joke:

\begin{quote}
What kind of ship can clean dishes whilst caring for your hands? {\em Ferry
liquid.}\footnote{Hand care is an advertised feature of ``Fairy Liquid''.}
\end{quote}
received the comment ``[This joke] lost points for saying `what kind of
ship,' and then having an answer that wasn't a ship,'', and an average
score of 1 point. The corresponding
reversed template with better word ordering produced the following joke
with the same schema and noun phrase:
\begin{quote}
What kind of detergent can cross water? {\em Ferry liquid.}
\end{quote}
is still not brilliant, but it is more logically coherent, and received
an average score of 2 points.

Only one template, (``what do you use to .... ....?'') was uniformly poor.
This is largely because the order of the lexemes is
of particular importance in questions of that form.
The template which puts the items in the reverse
order did quite well.

\subsection{Schema-Template Pairs}

If we were to eliminate the schemata and templates that always
produced poor jokes, the jokes would undoubtedly improve; however, we
would still be left with a lot of poor jokes. A large number of these
can be removed by severing the link between certain schemata and
templates, so that no jokes will be produced that use both.  However,
it is difficult to decide which schema-template links to sever.
Blindly removing all pairings which produce below-average mean scores
would indeed improve overall performance, but would not have a great
deal of theoretical significance, especially since some
schema-template pairs produced only one joke. Nonetheless, the pairs
can be put in order, in terms of average success, then investigated
further, to see if there is a sound reason to keep or eliminate any
particular pairings.

\subsection{Post-Production Checking}

Although none of the numbers can be interpreted as supporting the
hypotheses on post-production checking (see section~\ref{hypotheses}),
several of the general comments do:

\begin{description}

\item[question length:]  ``[The criteria used to judge the jokes were]
in order of importance: gut reaction, cleverness, delivery, rhythm. The
rhythm and delivery [of these jokes] could be improved. They have to be
`bang-bang' jokes.''

\item[alliteration and rhyming:] ``Phrasing is important. Numbers aren't
funny, but `quirky quantifier' is.'' This comment was referring to the
joke:
\begin{quote}
What do you call a quirky quantifier? {\em An odd number.}
\end{quote}

\item[`funny letters':]  ``The grizzly bare joke [What do you
call a ferocious nude? {\em A grizzly bare}] makes good use of the
funny letter Z.''

\item[subject matter:] ``I don't find `gay' jokes funny.'' This comment
was referring to the joke:

\begin{quote}
What do you call a homosexual that cleans dishes whilst caring for your
hands? {\em Fairy liquid.}
\end{quote}

\item[accidental associations:] ``Keep the ones with the double
meanings, eg [the one about] a gay being drunk.'' This was with
reference to the joke:

\begin{quote}
What do you call a homosexual you can drink? {\em Fairy liquid}.
\end{quote}
The judge who made this comment gave the joke in question a five and
commented on the ``obscene'' image it brought to mind, while the other
judge for this joke gave it a zero, possibly due to its subject matter.
\end{description}

\subsection{Miscellaneous comments}

Two factors were mentioned several times as being reasons for low
scores:
`joke fatigue', from reading too many similar jokes,  and a lack
of enthusiasm for this genre of joke.

\subsection{Jokes heard before}
\label{heard}

Perhaps one of the most heartening results of this evaluation is that
several of the judges claimed to have heard similar or identical jokes
before. This can be taken as evidence that \jape\ is producing some good
examples of the punning riddle genre.

Four judges claimed to had heard these \jape\ jokes, all variations on the
``cereal killer'' pun:

\begin{quote}

What do you call a murderer that has fibre? {\em A cereal killer.} \\
What do you get when you cross grain with a murderer? {\em A cereal
killer.}\\
What do you call breakfast food that murders people? {\em A cereal killer.}\\

\end{quote}

Other \jape\ jokes that had been heard before included:

\begin{quote}

What kind of tree can you wear? {\em A fir coat.}\\
What kind of rain has presents? {\em A bridal shower.}\\
What do you call a good-looking taxi? {\em A hansom cab.}\\
What do you call a perforated relic? {\em A holey grail.}\\
What do you get when you cross a savage pig with a drilling tool? {\em A
wild bore.}\\
\end{quote}

Another interesting point was that, well after the evaluation had taken
place, it was noticed that \jape\ had produced two jokes very similar to
ones in the Crack-a-Joke Book. \jape's jokes were:

\begin{quote}

What kind of pig can you ignore at a party? {\em A wild bore.}

\end{quote}
and
\begin{quote}
What do you use to help a lemon?\footnote{This riddle came up during
\jape's development, not during the evaluation.} {\em Lemon aid.}
\end{quote}
whereas the Crack-a-Joke versions were:
\begin{quote}
What runs around forests making other animals yawn? {\em A wild bore.}
\end{quote}
and
\begin{quote}
What do you give a hurt lemon? {\em Give it lemonade, of course.}
\end{quote}

\subsection{Post--evaluation trimming}

The heuristics suggested in section~\ref{hypotheses}, and confirmed by
the results, can be divided into two types: those that could
be implemented simply by removing sections of program in \jape, and
those that would require more programming or even a revision of the
design.

The former category includes the following:

\begin{itemize}

\item Removing the {\em double} schema;
\item Removing the poor templates;
\item Keeping the {\em elan} schema paired only with its more successful
templates, and
\item Removing lexical entries which do not meet the specifications
outlined in section~\ref{hypotheses}.

\end{itemize}

The results of all of these adjustments (save the last, which would
require the attentions of another common--knowledge judge) can be seen
immediately, simply by crossing out the jokes they would eliminate, and
reassessing the results.
When this is done, the average overall score of the jokes \jape\ produced
from the volunteered data rises to 1.9 points (``Not so bad'').
This is a large
improvement, especially considering the only changes to the program
itself were a few simple deletions; however, it suggests that
simply eliminating poor templates and schemata is just a first step
towards improving \jape's performance.

\subsection{Conclusions}

\label{improvements}

This evaluation has accomplished two things. It has shown that \jape\ can
produce pieces of text that are recognizably jokes (if not very good
ones) from a relatively unbiased lexicon. More importantly, it has
suggested some ways that \jape\ could be improved:
\begin{itemize}
\item The description of the lexicon could be made more precise, so that
it is easier for people unfamiliar with \jape\ to make appropriate entries.
Moreover, multiple versions of an entry could be compared for `common
knowledge', and that common knowledge entered in the lexicon.
\item More slots could be added to the lexicon, allowing the person
entering words to specify what a thing is made of, what it uses, and/or
what it is part of.
\item New, more detailed templates could be made so that the {\em
double\_pun} schema would produce comprehensible jokes.
\item Templates which use the lexemes given to them in the `wrong'
order (i.e.\ an order that suggests the words in the punchline should be
reversed) could be removed.
\item The remaining templates could be adjusted so that they use the
lexical data more gracefully, by providing the right amount of
information in the question part of the riddle.
\item Schema-template links that give consistently poor results
could be removed.
\end{itemize}

It is not easy to see how some of the other suggestions for improvement
could be implemented in \jape\ as it stands. However, these ideas could be
incorporated into later, more complex systems:
\begin{itemize}
\item Improve the delivery and rhythm of the jokes.
\item Attempt to make the punchline alliterate or rhyme when possible.
\item Maximize the number of `funny letters'.
\item Stick to inherently funny subject matter.
\item Generate jokes, then test them for serendipitous associations.
\end{itemize}

If even the simplest of the trimming and ordering heuristics described
above were implemented, \jape's output would be restricted to
good--quality punning riddles.

\section{Related work}
\label{others}

\subsection{The General Theory of Verbal Humour}

Although many studies have been done on the language of humour (for
example, \cite{Chiaro} and \cite{Booth}), few have looked at the
linguistics of humour in any depth. This is not to denigrate the work
that has been done in the field; however, in order for the computer
generation of riddles to be a tractable problem for symbolic AI, a
detailed, formal, linguistic model of (at least a subset of) humour is
required at some point.
The only model we found that even approaches the level of formality
required is Salvatore Attardo and Victor Raskin's General Theory of
Verbal Humour (GTVH) as described and developed in \cite{AR1},
\cite{AR2} and \cite{AR3}.

The General Theory of Verbal Humour is an attempt by Attardo and Raskin
to build a linguistically sound model of verbal
humour\footnote{This section is essentially a precis of \cite{AR1}.
Their examples are used.}. By analysing the similarities and differences
of a set of variants on a light-bulb joke, Attardo and Raskin postulate
six joke parameters, or {\em knowledge resources} (KR), which between them
determine the final text form of the joke. These KRs are organized into
a hierarchy.


\paragraph{Script Opposition:}
The {\em script opposition} KR is based on Raskin's earlier script-based
semantic theory of humour (SSTH), which he summarizes as follows:
\begin{quote}

A chunk of structured semantic information, the script can be
understood for the purposes of this article as an interpretation of the
text of a joke. The main claim of SSTH is that the text of a joke is
always fully or in part compatible with two distinct scripts and that
the two scripts are opposed to each other in a special way.
\cite{AR1}

\end{quote}

The `special ways' in which scripts can be opposed are at various levels
of abstraction, i.e., real vs. unreal, good vs. bad, high stature vs. low
stature, nondumb vs. dumb, etc. For example, the joke:

\begin{quote}
JOKE 1: How many Poles does it take to screw in a light bulb? Five. One to
hold the bulb and four to turn the table he's standing on. \cite{FH}
\end{quote}
uses the nondumb vs.. dumb opposition, since it is about applying a stupid
method to a simple task which most people deal with in a simple,
intelligent fashion.

\paragraph{Logical Mechanism:}
This parameter determines the mechanism used to oppose the scripts. For
example, joke 1 uses figure-ground reversal (from gestalt psychology,
according to Attardo and Raskin). Usually, when screwing in a light
bulb, the room, the ladder or table, the person screwing in the bulb,
etc. stay still, while the light bulb moves; joke 1  reverses that
situation.

Holding the other parameters of joke 1 constant, but changing the
logical mechanism to `false analogy', we get:

\begin{quote}
How many Poles does it take to screw in a light bulb? Five. One to
hold the light bulb and four to look for the right screwdriver.
\end{quote}

Other such mechanisms include simple reversal,  false priming, simple
juxtaposition, and ``the juxtaposition of two different situations
determined by the ambiguity or homonymy in a pun'' \cite{AR1}.

Note that, in the ``joke telling mode of communication'', the truth of
statements and their consistency become less important. The pseudologic
of the joke, therefore, need not be valid, just vaguely persuasive ---
persuasive enough that the listener will go along with the joke.

\paragraph{Situation:}
The situation of a joke is the set of details (e.g.\ time, place, objects,
activity, etc) which specify the joke. A given script opposition and
logical mechanism can be applied to a number of different situations.
For example,

\begin{quote}
 How many Poles does it take to wash a car? Two. One to hold the
sponge and one to move the car back and forth.
\end{quote}
differs from joke 1 only in situation.

\paragraph{Target:}
The target of a joke, the person or stereotype the joke is aimed at, is
the only optional parameter of the six. Many jokes have no identifiable
target. The target of joke 1 is, clearly, Poles, but it could be changed
to almost any other group which is stereotyped as `stupid' (e.g.\ blondes,
Newfies, etc).

\paragraph{Narrative Strategy:}
This KR determines the form the joke will take, i.e.\ riddle, conundrum,
expository text, etc. The more standard strategies have the advantage
that the punchline pretty much automatically falls in the right place.
Also, the choice of logical mechanism limits the range of narrative
strategies available.

Joke 1 as expository text, rather than conundrum, might look like this:

\begin{quote}
It takes five Poles to screw in a light bulb: one to hold the light
bulb and four to turn the table he's standing on.
\end{quote}

\paragraph{Language:}
This parameter specifies which paraphrasing of the joke is used (i.e.\ what
the surface form of the joke is). It is constrained by all the other
parameters. For example, although the language parameter determines the
exact phrasing and placement of the punchline, all the other parameters
(particularly narrative strategy and logical mechanism) have a lot of
input into it as well.

\subsubsection*{Relevance of GTVH to this project}
\label{theoryAR}
By providing a parameterized model of verbal humour in general, Attardo
and Raskin have provided a rough structure which could, in part, guide
the design of a humour producing program.

In particular, they note that holding some of their parameters constant
produces a joke `template'\footnote{Attardo and Raskin use this
term in a different way from the particular definition used in our
model of riddles.
}.
If we choose a few of the better defined
parameters to keep variable, and hold the rest constant, we should have
a constrained model of (certain types of) humour which could, in theory
at least, be used as the basis for a program design.

Unfortunately, their model is neither detailed nor formal enough to be
implemented as it stands, even in a constrained, `template' form. Some
of their ``knowledge resources'', in particular the script-opposition
and logical mechanism KRs, require a near-complete understanding of the
world (including the rules of physics, the operations of human society,
and common-sense reasoning) in order to operate. Even their language KR
``includes all the choices at the phonetic, phonologic, morphophonemic,
morphologic, lexic, syntactic, semantic, and pragmatic levels of
language structure that the speaker is still free to make'' as well as
``a few specifically humorous elements and relations'' \cite{AR1}.
In order for computer implementation of this model to be feasible,
it must be severely constrained,
perhaps so much as to be unrecognizable.

Nevertheless, our model could be seen as a shell
produced by constraining or holding
constant some of Attardo and Raskin's parameters. In terms of their six
knowledge resources, our model could be described as follows:

\begin{description}

\item[Language] The surface form of the joke is determined by the
template. The choice of template  is constrained by the choice of
schema, and the instantiation of the schema provides the template with
information it needs to construct the surface form.

\item[Narrative Strategy] The narrative strategy is fixed, since these
jokes are all, by definition, question-answer riddles.

\item[Target] There is no explicit target in any of the set of aimed-for
riddles, although it is possible that some object of ridicule might be
accidentally incorporated. If the lexicon were adjusted to include
stereotypical (instead of typical) semantic links, then jokes with butts
would probably arise more often.

\item[Logical Mechanism] The logical mechanism is fixed for noun-phrase
punning riddles: the juxtaposition of the meaning of a phrase, with the
meaning of a word which puns part of the phrase. The different
(uninstantiated) schemata could be seen as slight variations on this
basic mechanism. Other riddle types would have different, but fixed,
logical mechanisms. For example, a spoonerism riddle's mechanism could
be described as the juxtaposition of the meaning of a phrase with its
spoonerism.

\item[Situation] The situation is variable, though `choosing the
situation' really means choosing words/concepts which are consistent
with the logical mechanism (i.e.\ instantiating the schema).

\item[Script Opposition]
In our class of riddles, there is no strong evidence for the two scripts being
{\em opposed} --- they need only be different.
If there is any script opposition at all in
punning riddles, it is fixed, real vs. unreal.

\end{description}

The model described in section~\ref{model}, then, is consistent with
Attardo and Raskin's GTVH, although
most of their parameters are either fixed or constrained. The
substantive features of our model correspond to three  GTVH parameters:
an uninstantiated
schema corresponds to a particular {\em logical mechanism};
instantiating the schema corresponds to choosing a {\em situation};  and
filling a template produces the surface-form {\em language} of a joke.
However, our model does not emphasize the same features of joke structure
as Attardo and Raskin's theory. In particular,
Attardo and Raskin hold that script opposition is absolutely essential
for a piece of text to be considered a joke; whereas it is not even
represented in our model. Either script opposition is not actually very
important in question-answer punning riddles, or one script opposition
is implicit in and fixed for all such jokes.

\subsection{Other Joke Generators}

We are aware of very few other attempts to generate jokes using a
computer.

\subsubsection{The Light Bulb Joke Generator}
\label{libjog}

Attardo and Raskin have put together a simple joke generating program,
LIBJOG (Light Bulb JOke Generator) \cite{AR2}, mainly to show how poorly
simple cut-and-paste methods work. The first version combines an entry
for an commonly-stereotyped group, for example:

\begin{quote}
\begin{verbatim}
(i)(Poles ((activity1 hold the light bulb)
	   (numberX 1)
	   (activity2 turn the table he is standing on)
	   (numberY 4)))
\end{verbatim}
\end{quote}
with a template (in their sense of this term) for a light bulb joke:

\begin{quote}
How many (group name) does it take to screw in a light bulb? (NumberX).
One to (activity1) and (numberY) to (activity2). [Condition: $X = 1 + Y$.]
\end{quote}
to make, not surprisingly:

\begin{quote}
How many Poles does it take to screw in a light bulb? Five. One to hold
the light bulb and four to turn the table he's standing on.
\end{quote}

Although Attardo and Raskin claim that later versions of LIBJOG
``introduced more templates, more fields, and looser (and richer)
relations among them,'' they give no evidence of a significantly
improved method. The joke-generating mechanism seems to remain the same:
substitute the (humour-related) values in an entry for a
stereotyped group, directly into a light-bulb joke template like the one
above.

\jape\ differs from LIBJOG is several significant ways:

\begin{itemize}

\item \jape\ produces a wider range of riddles, which do not
have the fixed surface format of light-bulb jokes.

\item \jape\ is an implementation of a model of humour, albeit a very
simple one, rather than a program that can produce jokes in an
uninteresting way. Although \jape\ uses a lexicon and templates, there
is more to its method than simply pasting the two together.

\item \jape's lexicon is humour-independent, containing the kind of
information one might expect to find in a dictionary (albeit in a
different form); whereas LIBJOG's `lexicon' holds entries on
commonly-stereotyped groups (Poles, blondes, etc), with pieces of text
describing stereotypical ways of screwing in a light bulb.

\item \jape\ produces jokes which share deep traits, such as type of
ambiguity and relationships between key words; whereas
LIBJOG produces lightbulb jokes and only lightbulb jokes.

\end{itemize}

For the above reasons, \jape's successes and failures are of theoretical
interest, and improvements in \jape\ amount to refinements of a model of
(a subset of) humour. Improvements in LIBJOG, on the other hand, can
only be cosmetic, since it does not generate jokes in accordance with
any particular theory.

\subsubsection{De Palma and Weiner}

Some of the issues involved in jokes such as:

\begin{verse}
What has a mouth and cannot eat?\\

{\em A river.}

\end{verse}

\noindent
are discussed by \cite{DePalma}.  That work is very close in spirit
and intention to ours, with its concentration on formalising the
underlying semantic patterns in a very restricted subclass of punning
riddle, but their chosen type of riddle is very different from ours.
They have embodied the joke-creating knowledge in an algorithm which
searches a knowledge base for suitably related but homophonous items,
so it is not clear (from the relatively brief account) how they separate
(if at all)  knowledge about jokes from general knowledge.

They give no details of how much has been implemented and tested.

\section{Future directions}
\label{further-work}
\label{discussion}

\subsection{Overall evaluation of project}

We have shown that for one particular (non-trivial) subclass of
of question-answer punning riddles
it is possible to formulate a precise and computationally-tractable
symbolic model which describes a wide range of jokes in a systematic
way.

We have implemented a computer program (\jape) to produce jokes using this
model, and have tested this using an unbiased, humour-independent
lexicon whose contents were supplied by a set of volunteers.

We have informally evaluated the jokes generated in this way
by having them assessed by a set of human judges.

The results showed that the current version of \jape\ does generate good
jokes (including several that had been heard and recorded before).
Unfortunately, it also generates many bad jokes. Trimming the program, by
systematically removing poor lexical entries, templates, and schemata,
would improve the output significantly; adding filtering heuristics
would improve it even more. Also, in hindsight, we realized
that human-made punning riddles should have been mixed with \jape's
output to be judged, so that a fair comparison could have been made.

Although there is certainly room for improvement in \jape's performance,
it does produce recognizable jokes in accordance with a model of punning
riddles, which has not been done successfully by any other program we
know of. In that, it is a success.

The model that \jape\ uses was influenced by Pepicello and Green's
work on the language of riddles \cite{PG} and Attardo and Raskin's
General Theory of Verbal Humour \cite{AR1}, but it does not adhere strictly
enough to either work to be considered evidence for or against their
theories.

\subsection{Possible extensions and improvements}
\label{extensions}

There are three kinds of improvement that could be made to \jape:

\begin{itemize}

\item the present program could be trimmed,

\item heuristics could be added, to filter out poor jokes, and

\item the program could be extended to handle other joke types.

\end{itemize}

See Section~\ref{improvements} for a list of more detailed suggestions
along these lines.

Finally, \jape\ could be extended to handle other joke types. At present,
\jape\ can only generate question-answer punning riddles with noun-phrase
punchlines. One obvious direction of expansion would be to allow
different kinds of punchline, for example:

\begin{quote}

Why is it dangerous to play cards in the jungle? {\em Because there are
so many cheetahs.} \cite{CAJB}

\end{quote}
Although such jokes could be modelled with the same kind of
schema that we already have, the templates and the lexicon would have to
be much more complex {\em syntactically}, in order to generate
appropriate question-answer pairs.

Another possible direction of expansion would be to attempt to generate
simple spoonerisms and sub-word puns. Again, these could be modelled
with schemata similar to the present ones, albeit with different link
types (eg {\em rhymes\_with} instead of {\em homophone\_of}, for
spoonerisms). The templates would have to be different from, but not
more complex than, the templates \jape\ currently uses.

\subsection{Other directions for computer generated humour}

Even if all the extensions described in section~\ref{extensions} are
carried out, \jape\ will still only be able to generate phonologically
ambiguous question-answer riddles, which are just a small subset of
humorous texts.

It would be interesting to attempt to model and generate morphologically
and syntactically ambiguous riddles. For this to be successful, an
extensive investigation into linguistic ambiguity, and how it is used in
humour, would have to take place.

It would perhaps be easier to take what we have learned about
phonological ambiguity and apply it to other genres of humour. Story
jokes, for example, often have a punned aphorism (e.g.\ ``People who live
in grass houses shouldn't stow thrones'') as a punchline. Perhaps
current research into natural language story generation could, with some
more complex model of puns, produce such jokes.

\subsection*{Acknowledgements}

KB is grateful for financial help from
Canada Student Loans, the Overseas Research Students Scheme,
and the St Andrew's Society of Washington, DC.
We would like to thank Salvatore Attardo for his comments and
for giving us access to his unpublished work.


\bibliographystyle{alpha}
\bibliography{biblio}

@article{AR1,
   AUTHOR    = {Salvatore Attardo and Victor Raskin},
   TITLE     = {Script theory revis(it)ed: joke similarity and joke
       representation model},
   JOURNAL   = {Humor},
   YEAR      = 1991,
   NUMBER    = 3,
   PAGES     = {293--347},
   VOLUME    = 4
}

@article{AR2,
   AUTHOR    = {Salvatore Attardo and Victor Raskin},
   TITLE     = {Non-Literalness and Non-Bona-Fide in Language: Approaches to
       Formal and Computational Treatments of Humour and
       Irony},
   JOURNAL   = {Humor},
   YEAR      = {forthcoming 1994}
}

@book{Booth,
   AUTHOR    = {Booth, Wayne},
   TITLE     = {A Rhetoric of Irony},
   YEAR      = 1974,
   PUBLISHER = {University of Chicago Press}
}

@book{Bratko,
   AUTHOR    = {Bratko, Ivan},
   TITLE     = {Prolog Programming for Artificial Intelligence},
   YEAR      = 1990,
   PUBLISHER = {Addison-Wesley}
}

@incollection{Buchanan,
   AUTHOR    = {Buchanan, Bruce G.},
   TITLE     = {Artificial intelligence as an experimental science},
   BOOKTITLE = {Aspects of Artificial Intelligence},
   YEAR      = 1988,
   PUBLISHER = {Kluwer},
   EDITOR    = {Fetzer, J.H.},
   PAGES     = {209--250}

}

@book{Chiaro,
   AUTHOR    = {Chiaro, Delia},
   TITLE     = {The language of jokes: analysing verbal play},
   YEAR      = 1992,
   PUBLISHER = {Routledge},
   ADDRESS   =  "London"

}

@book{Chomsky57,
   AUTHOR    = {Chomsky, Noam},
   TITLE     = {Syntactic {S}tructures},
   YEAR      = 1957,
   PUBLISHER = {Mouton},
   ADDRESS =  {The Hague}
}

@book{Chomsky65,
   AUTHOR    = {Chomsky, Noam},
   TITLE     = {Aspects of the theory of Syntax},
   YEAR      = 1965,
   PUBLISHER = {MIT Press},
   ADDRESS =  {Cambridge, Mass.}
}

@book{FH,
   AUTHOR    = {Matt Freedman and Paul Hofman},
   TITLE     = {How Many Zen Buddhists Does It Take to Screw in a Light
       Bulb?},
   YEAR      = 1980,
   PUBLISHER = {St Martin's}
}

@incollection{Minsky2,
   AUTHOR    = {Minsky, Marvin},
   TITLE     = {Steps towards Artificial Intelligence},
   BOOKTITLE = {Computers and Thought},
   YEAR      = 1963,
   PUBLISHER = {McGraw-Hill},
   EDITOR    = {Feigenbaum, E.A. and Feldman, J.},
   PAGES     = {406--450}
}

@article{Newell,
   AUTHOR    = {Newell, Allan and Simon, Herbert},
   TITLE     = {Computer science as an empirical inquiry: symbols and search},
   JOURNAL   = {Communications of the ACM},
   YEAR      = 1976,
   PAGES     = {113--126},
   VOLUME    = 19
}

@inproceedings{DePalma,
   AUTHOR    = {De Palma, Paul and E. Judith Weiner},
   TITLE     = {Riddles: accessibility and knowledge representation},
   BOOKTITLE = {Proceedings of the 15th International Conference on
       Computational  Linguistics (COLING-92)},
   YEAR      = 1992,
   PAGES     = {1121--1125},
   VOLUME    = 4
}

@book{PG,
   AUTHOR    = {Pepicello, W.J. and Thomas A. Green},
   TITLE     = {The Language of Riddles},
   YEAR      = 1984,
   PUBLISHER = {Ohio State University}
}

@incollection{Ritchie,
   AUTHOR    = {Ritchie, Graeme},
   TITLE     = {Learning from {AM}},
   BOOKTITLE = {Artificial Intelligence in Mathematics},
   YEAR      = 1994,
   PUBLISHER = {Oxford University Press},
   EDITOR    = {Johnson, J.H. and McKee, S. and Vella, A.},
   ADDRESS   = {Oxford},
   PAGES     = {55--66}
}

@article{AR3,
   AUTHOR    = {Ruch, Wilhelm and Salvatore Attardo and Victor Raskin},
   TITLE     = {Toward an empirical verification of the General Theory of
       Verbal Humour},
   JOURNAL   = {Humor},
   YEAR      = 1993,
   NUMBER    = 2,
   VOLUME    = 6
}

@book{Townsend,
   AUTHOR    = {Townsend, William and Antworth, Evan},
   TITLE     = {Handbook of Homophones (online version)},
   YEAR      = 1993
}

@book{CAJB,
   TITLE     = {The Crack-a-Joke Book},
   YEAR      = 1978,
   PUBLISHER = {Puffin},
   EDITOR    = {Webb, Kaye}
}

\appendix
\newpage
\section{Summary of results}
\label{full_result_summary}

\subsection{Schemata Scores}

\begin{tabular}{|l|l|l|}
\hline
Aspect & Number of Jokes & Average Score \\
\hline\hline
{\em Schemata} & & \\ \hline
{\sc elan} & 31 & 1.3 \\ \hline
{\sc jumper} & 30 & 1.6 \\ \hline
{\sc lotus} & 42 & 1.7 \\ \hline
{\sc woolly} & 63 & 1.7 \\ \hline
{\sc double} & 22 & 1.1 \\ \hline
{\sc ginger} & 0 & 0 \\ \hline
{\sc Total} & 188 & 1.5 \\ \hline\hline
\end{tabular}

\subsection{Template Scores}

\begin{tabular}{|l|l|l|}
\hline
Aspect & Number of Jokes & Average Score \\
\hline\hline
{\em Templates } & & \\ \hline
{\sc syn\_syn} & 22 & 1.8 \\ \hline
{\sc syn\_verb} & 30 & 1.6 \\ \hline
{\sc syn\_verb\_rev} & 17 & 1.3 \\ \hline
{\sc use\_syn} & 4 & 0.6 \\ \hline
{\sc use\_syn\_rev} & 14 & 1.6 \\ \hline
{\sc class\_verb} & 25 & 1.4 \\ \hline
{\sc class\_verb\_rev} & 15 & 1.7 \\ \hline
{\sc class\_has} & 8 & 1.1 \\ \hline
{\sc class\_has\_rev} & 13 & 1.4 \\ \hline
{\sc adj\_syn} & 30 & 1.8 \\ \hline
{\sc adj\_syn\_rev} & 10 & 1.1 \\ \hline
{\sc Total} & 188 & 1.5 \\ \hline\hline
\end{tabular}

\subsection{Schema-Template Pair Scores}

\begin{tabular}{|l|l|l|}
\hline
Aspect & Number of Jokes & Average Score \\
\hline\hline
{\em Schema-Template Pairings } & & \\ \hline
{\sc elan + syn\_syn} & 3 & 1.7 \\ \hline
{\sc elan + syn\_verb} & 2 & 1.4 \\ \hline
{\sc elan + syn\_verb\_rev} & 5 & 1.4 \\ \hline
{\sc elan + use\_syn} & 4 & 0.6 \\ \hline
{\sc elan + use\_syn\_rev} & 1 & 1 \\ \hline
{\sc elan + class\_verb} & 4 & 1.2 \\ \hline
{\sc elan + class\_verb\_rev} & 4 & 1.8 \\ \hline
{\sc elan + class\_has} & 2 & 0.6 \\ \hline
{\sc elan + class\_has\_rev} & 2 & 1.5 \\ \hline
{\sc elan + adj\_syn} & 2 & 2.3 \\ \hline
{\sc elan + adj\_syn\_rev} & 2 & 0.8 \\ \hline
{\sc lotus + syn\_syn} & 2 & 1.8 \\ \hline
{\sc lotus + syn\_verb} & 9 & 1.5\\ \hline
{\sc lotus + syn\_verb\_rev} & 1 & 2.7 \\ \hline
{\sc lotus + use\_syn} & 0 & 0 \\ \hline
{\sc lotus + use\_syn\_rev} & 6 & 1.7 \\ \hline
{\sc lotus + class\_verb} & 8 & 1.5 \\ \hline
{\sc lotus + class\_verb\_rev} & 4 & 1.9 \\ \hline
{\sc lotus + class\_has} & 2 & 0.8 \\ \hline
{\sc lotus + class\_has\_rev} & 2 & 2.0 \\ \hline
{\sc lotus + adj\_syn} & 6 & 2.4 \\ \hline
{\sc lotus + adj\_syn\_rev} & 2 & 0.4 \\ \hline
{\sc woolly + syn\_syn} & 9 & 2.1 \\ \hline
{\sc woolly + syn\_verb} & 13 & 2.0 \\ \hline
{\sc woolly + syn\_verb\_rev} & 7 & 1.2 \\ \hline
{\sc woolly + use\_syn} & 0 & 0 \\ \hline
{\sc woolly + use\_syn\_rev} & 2 & 1.0 \\ \hline
{\sc woolly + class\_verb} & 6 & 1.9 \\ \hline
{\sc woolly + class\_verb\_rev} & 3 & 2.0 \\ \hline
{\sc woolly + class\_has} &  & 1.0 \\ \hline
{\sc woolly + class\_has\_rev} & 5 & 1.3 \\ \hline
{\sc woolly + adj\_syn} & 13 & 1.6 \\ \hline
{\sc woolly + adj\_syn\_rev} & 3 & 1.2 \\ \hline
{\sc jumper + syn\_syn} & 6 & 1.8 \\ \hline
{\sc jumper + syn\_verb} & 3 & 1.2 \\ \hline
{\sc jumper + syn\_verb\_rev} & 2 & 1.0 \\ \hline
{\sc jumper + use\_syn} & 0 & 0 \\ \hline
{\sc jumper + use\_syn\_rev} & 2 & 2.4 \\ \hline
{\sc jumper + class\_verb} & 2 & 0.5 \\ \hline
{\sc jumper + class\_verb\_rev} & 3 & 1.2 \\ \hline
{\sc jumper + class\_has} & 1 & 3.3 \\ \hline
{\sc jumper + class\_has\_rev} & 2 & 1.8 \\ \hline
{\sc jumper + adj\_syn} & 8 & 1.5 \\ \hline
{\sc jumper + adj\_syn\_rev} & 1 & 2.5 \\ \hline
{\em cont...} & & \\ \hline
\end{tabular}

\begin{tabular}{|l|l|l|}
\hline
Aspect & Number of Jokes & Average Score \\
\hline\hline
{\em Schema-Template Pairings cont.} & & \\ \hline
{\sc double + syn\_syn} & 2 & 1.2 \\ \hline
{\sc double + syn\_verb} & 3 & 1.1 \\ \hline
{\sc double + syn\_verb\_rev} & 2 & 0.5 \\ \hline
{\sc double + use\_syn} & 0 & 0 \\ \hline
{\sc double + use\_syn\_rev} & 3 & 1 \\ \hline
{\sc double + class\_verb} & 5 & 1.1 \\ \hline
{\sc double + class\_verb\_rev} & 1 & 1.3 \\ \hline
{\sc double + class\_has} & 1 & 0 \\ \hline
{\sc double + class\_has\_rev} & 2 & 0.7 \\ \hline
{\sc double + adj\_syn} & 1 & 1 \\ \hline
{\sc double + adj\_syn\_rev} & 2 & 2 \\ \hline
\end{tabular}

\subsection{Phrase Scores}

\begin{tabular}{|l|l|l|}
\hline
Aspect & Number of Jokes & Average Score \\
\hline\hline
{\em Noun Phrases } & & \\ \hline
{\sc aeroplane hangar} & 4 & 1.0 \\ \hline
{\sc bridal shower} & 16 & 1.1 \\ \hline
{\sc bridle path} & 2 & 0.8 \\ \hline
{\sc broken hart} & 4 & 2.1 \\ \hline
{\sc coral reef} & 1 & 1.0 \\ \hline
{\sc dual carriageway} & 7 & 1.8 \\ \hline
{\sc fairy liquid} & 23 & 1.5 \\ \hline
{\sc first base} & 1 & 1.0 \\ \hline
{\sc foul play} & 4 & 1.3 \\ \hline
{\sc fur coat} & 15 & 1.9 \\ \hline
{\sc grizzly bear} & 4 & 1.9 \\ \hline
{\sc hansom cab} & 4 & 2.0 \\ \hline
{\sc holy grail} & 2 & 1.7 \\ \hline
{\sc love bite} & 4 & 1.9 \\ \hline
{\sc odd number} & 8 & 2.7 \\ \hline
{\sc pool cue} & 33 & 1.2 \\ \hline
{\sc serial killer} & 10 & 2.6 \\ \hline
{\sc signet ring} & 13 & 0.9 \\ \hline
{\sc sole heir} & 16 & 1.6 \\ \hline
{\sc wild boar} & 17 & 1.4 \\ \hline
\end{tabular}

\end{document}